\documentclass[journal]{IEEEtran}
\pdfoutput=1 
\usepackage{amsmath}
\usepackage{amssymb}
\usepackage{ntheorem}
\usepackage{subfigure}
\usepackage{caption}
\usepackage{indentfirst}
\usepackage{ifpdf}
\usepackage{cite,color}
\usepackage{graphicx}
\usepackage{epstopdf}
\usepackage{array}
\usepackage{cite}
\usepackage{algorithm}
\usepackage{algorithmic}
\usepackage{multirow}
\usepackage{booktabs}
\usepackage{threeparttable}
\usepackage{diagbox}

\newcommand{\tabincell}[2]{\begin{tabular}{@{}#1@{}}#2\end{tabular}}

\newtheorem{theorem}{Theorem}

\begin{document}
\title{Hankel Matrix Nuclear Norm Regularized Tensor Completion for  $N$-dimensional Exponential Signals }
\author{Jiaxi~Ying,
        Hengfa~Lu,
        Qingtao~Wei,
        Jian-Feng~Cai,
        Di~Guo,
        Jihui~Wu,
        Zhong~Chen,
        Xiaobo~Qu*% <-this % stops a space
\thanks{This work was supported by National Natural Science Foundation of China (61571380, 61302174, 61201045 and 11375147), Natural Science Foundation of Fujian Province of China (2015J01346, 2016J05205), Fundamental Research Funds for the Central Universities (20720150109), Hong Kong Research Grant Council (16300616), Important Joint Research Project on Major Diseases of Xiamen City (3502Z20149032) and the Strategic Priority Research Program of the Chinese Academy of Sciences (XDB08030302).(*Corresponding author: Xiaobo Qu)}
\thanks{Jiaxi Ying, Hengfa Lu, Zhong Chen and Xiaobo Qu are with the
Department of Electronic Science, Fujian Provincial Key Laboratory of
Plasma and Magnetic Resonance, Xiamen University, Xiamen, China (e-mail:
 quxiaobo@xmu.edu.cn)}% <-this % stops a space
\thanks{Qingtao Wei and Jihui Wu are with School of Life Sciences,
University of Science and Technology of China, Hefei, Anhui, China.}
\thanks{Jian-Feng Cai is with Department of Mathematics, Hong Kong University
of Science and Technology, Hong Kong SAR, China. }
\thanks{Di Guo is with the School of Computer and Information Engineering,
Fujian Provincial University Key Laboratory of Internet of Things Application
Technology, Xiamen University of Technology, Xiamen, China.}}% <-this % stops a space

%===========================================================
%-------------------------- abstract -----------------------
%===========================================================
\maketitle
\begin{abstract}
Signals are generally modeled as a superposition of exponential functions in spectroscopy of chemistry, biology and medical imaging. For fast data acquisition or other inevitable reasons, however, only a small amount of samples may be acquired and thus how to recover the full signal becomes an active research topic. But existing approaches can not efficiently recover $N$-dimensional exponential signals with $N\geq 3$. In this paper, we study the problem of recovering $N$-dimensional (particularly $N\geq 3$) exponential signals from partial observations, and formulate this problem as a low-rank tensor completion problem with exponential factor vectors. The full signal is reconstructed by simultaneously exploiting the CANDECOMP/PARAFAC structure and the exponential structure of the associated factor vectors. The latter is promoted by minimizing an objective function involving the nuclear norm of Hankel matrices. Experimental results on simulated and real magnetic resonance spectroscopy data show that the proposed approach can successfully recover full signals from very limited samples and is robust to the estimated tensor rank.
\end{abstract}


%===========================================================
%------------------------ Introduction ---------------------
%===========================================================
\IEEEpeerreviewmaketitle
\section{Introduction}
\IEEEPARstart{S}{ignal} reconstruction from its sampled measurements is recognized as a fundamental theme of signal processing. Under some circumstances, these acquired measurements are incomplete due to costly experiments, hardware limitation, or other inevitable reasons. For example, for the purpose of fast data acquisition, nonuniform sampling is used to obtain partial entries of the time-domain signal in nuclear magnetic resonance (NMR) spectroscopy \cite{sue13, sue14, sue43}, which has been widely used in chemistry and biology. Recovering the full signal is essential for the next step of data analyses in these applications.

In this paper, the signal of interest $f( {{t_1},{t_2}, \ldots ,{t_N}})$ can be modeled or approximated by a superposition of limited $R$ $N$-dimensional ($N$-D) exponentials, i.e.,
\begin{equation}\label{eq:model}
f({{t_1},{t_2},\ldots,{t_N}}) = \sum\limits_{r = 1}^R {{d_r}{\phi _{1,r}}({{t_1}})}{\phi_{2,r}}({{t_2}}) \cdots {\phi_{N,r}}({{t_N}}),
\end{equation}
where $f$ denotes a scalar function of $N$ variables; ${\phi _{n,r}}\left( {{t_n}} \right)$ the exponential function for any $n$ ($1\leq n \leq N$) and $r$ ($1\leq r \leq R$), and ${d_r}$ the complex amplitudes of the associated coefficient. This formulation implies that each component $N$-D exponential in \eqref{eq:model} is an exponential function with respect to any variable. In particular, when ${\phi _{n,r}}\left( {{t_n}} \right) = {e^{j2\pi {f_{n,r}}{t_n}}}$, $f$ denotes a spectrally sparse signal \cite{sue15, sue16}, i.e., a superposition of $R$ $N$-D undamped complex sinusoids. When ${\phi _{n,r}}\left( {{t_n}} \right) = {e^{\left( { - 1/{\tau _{n,r}} + j2\pi {f_{n,r}}} \right){t_n}}}$ with the damping factor ${\tau _{n,r}}\in\mathbb{R}_+$, $f$ denotes a sum of $R$ $N$-D damped complex sinusoids. These two types of signals arise in various applications such as multiple-input multiple-output (MIMO) radars \cite{sue3}, harmonic analysis \cite{sue5,sue9, sue17}, analog-to-digital conversion \cite{sue10}, magnetic resonance imaging \cite{sue11, sue12}, and NMR spectroscopy \cite{sue13,sue63}. In this paper, we study the problem of reconstructing $N$-D exponential signals from a small amount of measurements.

\begin{figure*}[ht]
\centering
\includegraphics[width=6.0in]{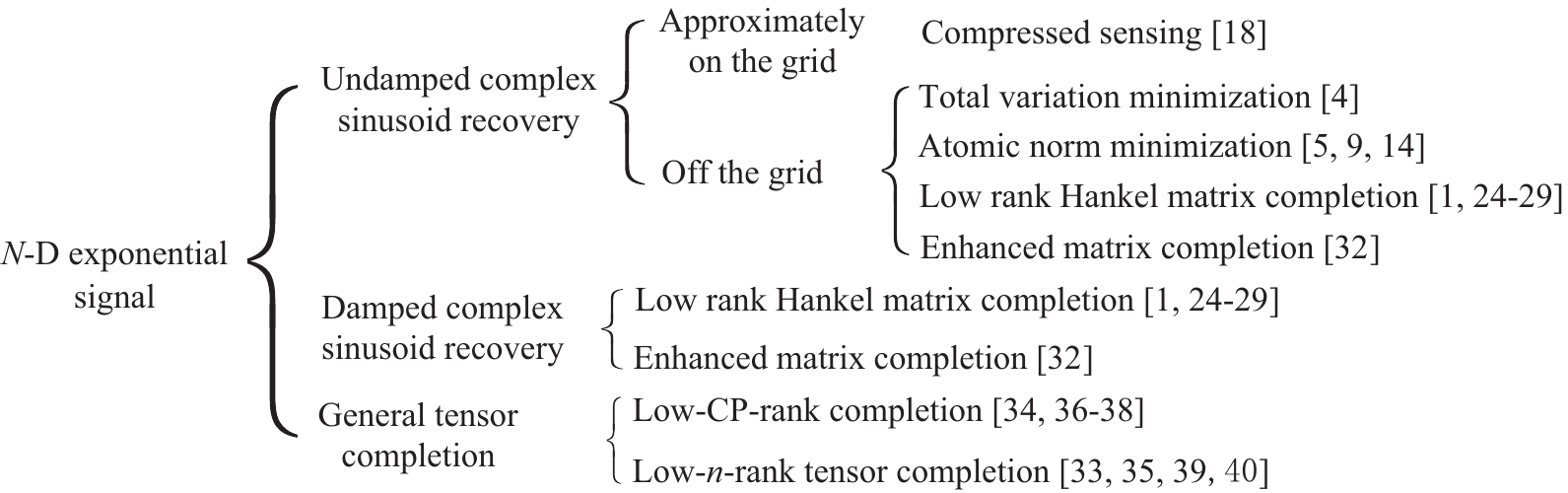}
\caption{Summary of methods in recovering \emph{N}-dimensional exponential signals. }
\label{Summary_of_Methods}
\end{figure*}

Recently, recovering a spectrally sparse signal becomes of great interest in signal processing community \cite{sue15,sue16,sue17,sue18,sue70,sue64,sue65}. Among emerging approaches, the compressed sensing \cite{sue19} suggests to reconstruct a signal from its partial observations if it enjoys a sparse representation in some transform domain and the observation operator satisfies some incoherence conditions. The spectrally sparse signal can be sparse in the discrete Fourier transform domain if the frequencies are aligned well with the discrete frequencies and the number of exponentials is small. In this case, signals can be recovered from very few measurements by enforcing the sparsity in the discrete Fourier domain \cite{sue19}. However, true frequencies in practical applications generally take values on a continuous domain, and the resultant basis mismatch between the true frequencies and the discretized grid \cite{sue20} leads to the loss of sparsity and hence worsens the performance of compressed sensing. For the purpose of addressing this problem, total variation or atomic norm \cite{sue21} minimization methods were proposed to deal with signal recovery with continuous-valued frequencies \cite{sue15, sue16, sue17, sue18}. However, these methods \cite{sue15,sue16,sue17, sue18} work only for undamped complex sinusoids but not for generic exponential signals, such as damped complex sinusoids. Furthermore, being computationally expensive, they search the solution in a space whose dimension is the \emph{square} of the dimension of the underlying signal, which is intractable for large scale problems (e.g., $N\geq 3$).

More recently, inspired by matrix pencil method \cite{sue22} and matrix completion \cite{sue23, sue24}, low rank Hankel matrix (LRHM) reconstruction \cite{sue13, sue26, sue27, sue28, sue25, sue29, Hankel_STSP} has been proposed to recover generic exponential signals. The LRHM was proposed in missing data recovery of non-uniformly sampling in realistic protein NMR spectroscopy \cite{sue13}, showing that broad peaks can be recovered much more reliably than minimizing the $\emph{l}_1$ norm on the spectrum \cite{sue14, sue43, sue59, sue60}. The minimal number of samples for LRHM method is theoretically predicted when the measurements are taken with random Gaussian sampling \cite{sue25}. However, LRHM is limited to recovering only 1-D signals. The $N$-D exponential signal is possibly recovered by enhanced matrix completion (EMaC) \cite{sue30} with theoretical guarantee of the stable recovery. Nonetheless, EMaC invokes minimization problems with a huge unknown matrix, prohibiting its applicability to $N$-D ($N\geq 3$) exponential signals. Note that the structured matrix based methods \cite{sue18,sue30} become prohibitive even in the 3-order tensor completion, because they solve the tensor completion problem by lifting to a search space whose dimension is on the order of the square of the dimension of the tensor. For example, for a signal of size ${50\times 50\times 50}$, EMaC \cite{sue30} needs to solve an optimization problem that involves a matrix of size $15625\times 17576$. Therefore, how to recover \emph{N}-D (particularly $N\geq3$) exponential signals still remains challenging.

It can be easily checked from \eqref{eq:model} that the signal of interest, viewed as an $N$-D tensor in the discrete domain, enjoys a low CANDECOMP/PARAFAC (CP) rank and a low $n$-rank if $R$ is sufficiently small. Therefore, the exponential signal recovery can be formulated as a low-rank tensor completion. The low-rank tensor completion has been successfully applied to a large class of real-world problems, such as data analyses in computer vision \cite{sue31, sue32}, remote sensing \cite{sue34}, and electroencephalogram \cite{sue33}. Since the rank of a tensor can be defined differently, there exist several types of low-rank tensor completion approaches, e.g.,  the low-CP-rank tensor completion \cite{sue32, sue33, sue35, sue41} and the low-$n$-rank tensor completion \cite{sue31, sue34, sue36, TSP_2013}. Although both low-CP-rank and low-$n$-rank tensor completions can be applied to recover \emph{N}-D (\emph{N}$\ge$3) signals, they ignore the specific exponential structure, e.g. time domain signal in NMR spectroscopy, probably causing the requirement of an unnecessarily large number of observed entries for a stable recovery.

In this paper, we propose an approach to reconstruct the $\textit{N}$-D exponential signal satisfying \eqref{eq:model} from a small amount of samples. The proposed approach simultaneously exploits the low-CP-rank structure and the exponential structure of the associated factor vectors (See the definition in Section \ref{sec:tensorcomp}). To enforce the former structure, we represent the signal in the CP decomposition form and imposed a least square fitting to the sampled data. To promote the latter structure, we penalize the nuclear norm of the Hankel matrices formed by factor vectors. We will verify, with comprehensive numerical experiments on simulated and real-world data, the effectiveness of the new method by comparisons with state-of-the-art tensor completion methods.

The rest of this paper is organized as follows. In Section \ref{sec:notations}, we will introduce the notations and related backgrounds. Section \ref{sec:tensorcomp} converts the $N$-D exponential signal reconstruction to a low-rank tensor completion problem and reviews existing low-rank tensor completion methods. In Section \ref{sec:HMRTC}, we will propose our Hankel matrix nuclear norm regularized tensor completion approach, including the algorithm, convergence and complexity analysis. Section \ref{sec:exp} presents the experimental results on simulated and real-world data. Section \ref{sec:discussion} discusses higher-dimensional experiments, comparisons with other state-of-the-art methods and some parameters. Section \ref{sec:conclusion} concludes this work and discusses the future work.

%===========================================================
%----------------- NOTATIONS AND BACKGROUNDS ---------------
%===========================================================
\section{NOTATIONS AND BACKGROUNDS}\label{sec:notations}
Notations and backgrounds of tensors are given below.

% -------------------------- Notations ---------------------
\subsection{Notations}
Notations and nomenclatures of tensors are introduced following the review paper \cite{sue37} and a summary of these notations is shown in Table \ref{tab:notations}.

Tensor is the generalization of matrix to high dimensions. For a tensor ${\cal X}\in\mathbb{C}^{I_{1}\times \cdots\times I_{N}}$, the number of dimensions $N$ is called the order, also known as way or mode. Throughout the paper, the $i$-th entry of a vector $\mathbf{x}$ is denoted by $x_{i}$, the $(i,j)$-element of a matrix $\mathbf{X}$ is denoted by $x_{i,j}$, and the $\left(i_1,i_2,\ldots,i_N\right)$-element of an order-$N$ tensor $\mathcal{X}$ is denoted by $x_{i_1,i_2,\ldots,i_N}$.

\begin{table}[!ht]
\renewcommand\arraystretch{1.5}
\caption{Basic notations}\label{tab:notations}
\centering
\begin{tabular}{cc}
\toprule
 Symbols & Notations \\
\midrule
$x,{\bf{x}},{\bf{X}}, {\cal X} $ & scalar, vector, matrix, tensor\\
${\bf{X}} = [{{\bf{x}}_1}{\rm{,}}{{\bf{x}}_2}{\rm{,}}  \cdots {\rm{,}} {{\bf{x}}_R}]$ & matrix $\bf{X}$ with column vectors ${\bf{x}_r}$ \\
${{\bf{X}}_{(n)}}$  & mode-\emph{n} matricization of tensor ${\cal X}$ \\
${{\bf{X}}^{T}}$,{\kern 1pt} ${{\bf{X}}^H}$&transpose, hermitian transpose\\
${{\bf{x}}_{{i_1}, \ldots, {i_{n - 1}},{\rm{:}},{i_{n + 1}}, \ldots, {i_N}}}$&mode-\emph{n} fiber of tensor ${\cal X}$\\
${{\bf{X}}_{{\bf{:}}, {\bf{:}}, {i_3}, \ldots, {i_N}}}$&matrix slice of tensor ${\cal X}$\\
\bottomrule
\end{tabular}
\end{table}

\setlength\parindent{0.5em} \paragraph{Fibers and Slices} Fibers are the higher-order analogue of matrix rows and columns. The fiber is a collection of entries of the tensor by fixing all but one indices. The mode-$n$ fibers are all vectors ${{\bf{x}}_{{i_1}, \ldots,  {i_{n - 1}},{\rm{:}},{i_{n + 1}}, \ldots, {i_N}}}$ that are obtained by fixing the indices $\left\{ {{i_1}, \ldots ,{i_N}} \right\}\backslash {i_n}$. Slices are $2$-D sections of a tensor, defined by fixing all but two indices.

\paragraph{Kronecker product} The Kronecker product of matrices ${\bf{A}} \in {\mathbb{C}^{m \times n}}$ and ${\bf{B}} \in {\mathbb{C}^{p \times q}}$, denoted by ${\bf{A}} \otimes {\bf{B}}$, is defined by
\begin{equation}
{\bf{A}} \otimes {\bf{B}}: = {[{a_{ij}}{\bf{B}}]_{mp \times nq}}
\end{equation}

\paragraph{Khatri-Rao product} The Khatri-Rao product is the columnwise Kronecker product. Given two matrices ${\bf{A}} \in {\mathbb{C}^{p \times n}}$ and ${\bf{B}} \in {\mathbb{C}^{q \times n}}$, their Khatri-Rao product ${\bf{A}} \odot {\bf{B}}$  satisfies
\begin{equation}
{\bf{A}} \odot {\bf{B}} = \left[ {\begin{array}{*{20}{c}}
{{{\bf{a}}_1} \otimes {{\bf{b}}_1}}&{{{\bf{a}}_2} \otimes {{\bf{b}}_2}}& \cdots &{{{\bf{a}}_n} \otimes {{\bf{b}}_n}}
\end{array}} \right]
\end{equation}
where $\otimes$ denotes Kronecker product.

\paragraph{Frobenius norm} The Frobenius norm of a tensor is the square root of the sum of the squares of the absolute value of each element, i.e.,
\begin{equation}
{\left\| {\cal X} \right\|_F} = {\biggl( {\sum\limits_{{i_1} = 1}^{{I_1}} {\sum\limits_{{i_2} = 1}^{{I_2}} { \cdots \sum\limits_{{i_N} = 1}^{{I_N}} \left|{x_{{i_1},  \cdots,{i_N}}} \right|^2 } } } \biggr)^{\frac{1}{2}}}{\rm{.}}
\end{equation}

% ---------- CP decomposition and tensor CP rank -----------
\subsection{CP decomposition and tensor CP-rank}
\paragraph{factor and rank-1 tensor}
An $N$-order tensor ${\cal X}$ is rank-$1$ if it can be written as the outer product of vectors as follows
$$
{\cal X} = {{\bf{u}}^{( 1 )}} \circ {{\bf{u}}^{( 2 )}} \circ  \cdots  \circ {{\bf{u}}^{( N )}},
$$
where the symbol $ \circ $ denotes the vector outer product and the vector ${{\bf{u}}^{\left( n \right)}}$, for all $1\leq n\leq N$, is called a \emph{factor}. This means each element of the tensor is the product of the corresponding vector elements:
$$
x_{{i_1}, \cdots,{i_N}} = \prod\limits_{n = 1}^N {u_{{i_n}}^{\left( n \right)}}
$$
for all $1 \le {i_n} \le {I_n}$ and $1\leq n\leq N$.

\paragraph{CP decomposition and tensor CP-rank}
The CP decomposition \cite{sue38}  factorizes a tensor into a linear combination of rank-$1$ tensors. A tensor ${\cal X} \in {\mathbb{C}^{{I_1} \times  \cdots  \times {I_N}}}$  is represented with CP decomposition as

\begin{equation}\label{eq:CPDec}
{\cal X} = \sum\limits_{r = 1}^R {{d_r}{\bf{u}}_r^{\left( 1 \right)} \circ {\bf{u}}_r^{\left( 2 \right)} \circ  \cdots  \circ {\bf{u}}_r^{\left( N \right)}},
\end{equation}
where $R$ is a positive integer and $d_r\in\mathbb{C}$. The tensor CP-rank is defined as the smallest number of rank-$1$ tensors that composes the ${\cal X}$ in \eqref{eq:CPDec}.

Following \cite{sue39}, we combine the factors in CP decomposition \eqref{eq:CPDec} to form \emph{factor matrices}
$$
{\bf{U}}^{( n )} = \bigl[
{\bf{u}}_1^{( n )} {\bf{u}}_2^{( n )} \ldots {\bf{u}}_R^{(n)}\bigr]
$$
for $1\leq n\leq N$, and the CP decomposition \eqref{eq:CPDec} is concisely expressed as
\begin{equation*}
\begin{split}
\mathcal{X}=
\bigl[\kern-0.25em\bigl[
{\bf{d}};{\bf{U}}^{(1)},{\bf{U}}^{(2)},\ldots,{\bf{U}}^{(N)}
\bigr]\kern-0.25em\bigr]
\end{split}
\end{equation*}
where ${\bf{d}}=[d_1,\ldots,d_R]^T \in {\mathbb{C}^R}$ and $[\kern-0.15em[ {}~]\kern-0.15em]$ is called the Tucker operator \cite{sue39}. Note that we can always rescale the factor matrices so that all the entries in $\mathbf{d}$ are 1's, and for simplicity, we will drop $\mathbf{d}$, i.e.,
\begin{equation*}
\bigl[\kern-0.25em\bigl[{\bf{U}}^{(1)},{\bf{U}}^{(2)},\ldots,{\bf{U}}^{(N)}
 \bigr]\kern-0.25em\bigr]
 \equiv
 \sum\limits_{r = 1}^R {{\bf{u}}_r^{\left( 1 \right)} \circ {\bf{u}}_r^{\left( 2 \right)} \circ  \cdots  \circ {\bf{u}}_r^{\left( N \right)}}.
\end{equation*}

% --------- Tensor matricization and Tensor n-rank ----------
\subsection{Tensor matricization and Tensor n-rank}
There are some other definitions of tensor rank. The $n$-rank is defined via tensor \emph{matricization} (also known as \emph{unfolding}), reordering the elements of a tensor into a matrix. The mode-\emph{n} matricization of a tensor ${\cal X}\in\mathbb{C}^{I_{1}\times \cdots\times I_{N}}$  is denoted by ${{\bf{X}}_{\left( n \right)}}$, and the tensor element $x_{{i_1},\cdots,{i_N}}$ is mapped to the matrix element $\left( {{i_n},j_n} \right)$, where
\begin{equation}
j_n = 1 + \sum\limits_{\scriptstyle k = 1,
\scriptstyle k \ne n\hfill}^N {\left( {{i_k} - 1} \right){J_k}{\kern 5pt} {\rm{with}}{\kern 5pt} {J_k} = \prod\limits_{\scriptstyle m = 1,
\scriptstyle m \ne n\hfill}^{k - 1} {{I_m}}}.
\end{equation}

Therefore, ${{\bf{X}}_{( n )}} \in {\mathbb{C}^{{I_n}\times{K_n}}}$ with ${K_n} = \prod_{k = 1,k \ne n}^N {{I_k}} $.
With the help of Khatri-Rao product, the mode-\emph{i} matricization can be expressed as
\begin{equation*}
\scalebox{0.95}{${{\bf{X}}_{\left(i\right)}}={{\bf{U}}^{\left(i\right)}}\left({{{\bf{U}}^{\left(N\right)}}\odot\cdots\odot{{\bf{U}}^{\left({i+1}\right)}}\odot{{\bf{U}}^{\left({i-1}\right)}}\odot\cdots\odot{{\bf{U}}^{\left(1 \right)}}}\right)^T$}.
\end{equation*}

The $n$-rank of an $N$-D tensor $\mathcal{X}$ is the tuple of the ranks of the mode-$n$ unfoldings:
\begin{equation*}
\scalebox{0.95}{${n\text{-rank}}\left({{\cal X}}\right)=\left({{\rm{rank}}\left({{{\bf{X}}_{\left(1\right)}}}\right),{\rm{rank}}\left({{{\bf{X}}_{\left(2\right)}}}\right), \ldots,{\rm{rank}}\left({{{\bf{X}}_{\left(N\right)}}}\right)}\right)$}.
\end{equation*}

%==============================================================================================
%----------------- N-D exponential signal reconstruction as a tensor completion ---------------
%==============================================================================================
\section{$N$-D exponential signal reconstruction as a tensor completion}\label{sec:tensorcomp}
\begin{figure}[htbp]
\centering
\includegraphics[width=3in]{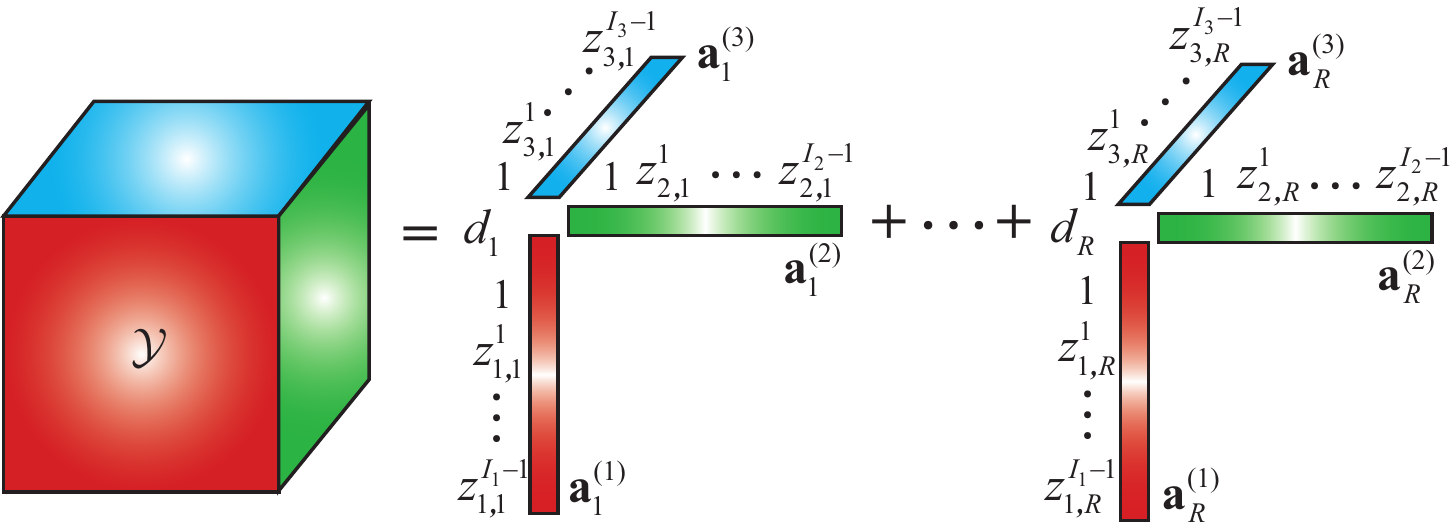}
\caption{CP decomposition for the signal of interest in 3-D.}
\label{CP_Decomposition}
\end{figure}

Without loss of generality, the frequencies in \eqref{eq:model} can be normalized with respect to the Nyquist frequency and hence the measurements are sampled at integer values. Therefore, by sampling the signal \eqref{eq:model} on a uniform grid, we can obtain an $N$-order tensor ${{\cal Y}} \in {\mathbb{C}^{{I_1} \times  \cdots  \times {I_N}}}$, and each entry ${y_{{i_1},\cdots,{i_N}}}$ can be expressed as
\begin{equation}\label{eq:signalmodel}
{y_{{i_1},\cdots,{i_N}}} = \sum\limits_{r = 1}^R \left({{d_r}\prod\limits_{n = 1}^N {z_{n,r}^{{i_n} - 1}} }\right),
\end{equation}
for ${i_n} = 0, \ldots ,{I_n}-1$, $n = 1{\rm{,}} \ldots {\rm{,}}N$, with ${d_r} \in \mathbb{C}$, and ${z_{n,r}} \in \mathbb{C}$. Fig. \ref{CP_Decomposition} shows a graphical illustration of \eqref{eq:signalmodel} when $N=3$, implying that each component, rank-$1$ tensor in $3$-D exponential signal \eqref{eq:signalmodel}, can be written as the outer product of vectors that are of exponential structure. Here we assume $R \ll \prod_{n = 1}^N {{I_n}} $.

In this paper, we aim to recover $\mathcal{Y}$ from its small subset of entries $\mathcal{P}_{\Omega}(\mathcal{Y})$,
\begin{equation}\label{eq:mea}
[\mathcal{P}_{\Omega}({\cal Y})]_{i_1,\ldots,i_N} =
\begin{cases}
y_{i_1,\ldots,i_N}&\mbox{if }(i_1,\ldots,i_N)\in\Omega,\cr
0&\mbox{otherwise}.
\end{cases}
\end{equation}

One can check that $\mathcal{Y}$ is low-rank in the sense of both CP-rank and $n$-rank if $R$ is sufficiently small. Therefore, the problem of reconstructing the signal $\mathcal{Y}$ can be recast as a low-rank \emph{tensor completion}. Depending on how the tensor rank is defined, we have a couple of possible $N$-D exponential signal reconstruction approaches using available low-rank tensor completion methods.

% ---------------------- Low CP-rank tensor completion ----------------
\subsection{Low-CP-rank tensor completion}
The signal $\mathcal{Y}$ enjoys a low CP-rank. Actually, \eqref{eq:signalmodel} implies the following CP-decomposition
\begin{equation}\label{eq:CPY}
{\cal Y} = \bigl[\kern-0.25em\bigl[{\bf{d}}; {\bf{A}}^{(1)},{\bf{A}}^{(2)},\ldots,{\bf{A}}^{(N)}
 \bigr]\kern-0.25em\bigr],
\end{equation}
where ${\bf{d}} = \left[ {{d_1}, \ldots ,{d_R}} \right]^T$ and the factor matrices ${{\bf{A}}^{\left( n \right)}} \in {\mathbb{C}^{{I_n} \times R}}$ are
\begin{equation}\label{eq:factormatrix}
{{\bf{A}}^{(n)}} = \bigl[ {{\bf{a}}_1^{(n)}, \ldots ,{\bf{a}}_R^{(n)}} \bigr]
\end{equation}
with
\begin{equation}\label{eq:factor}
{\bf{a}}_r^{\left( n \right)} = {\bigl[ {1,{\kern 1pt} {z_{n,r}},{\kern 1pt} z_{n,r}^2, \ldots ,z_{n,r}^{{I_n} - 1}} \bigr]^T}.
\end{equation}
These equations imply that $\mathcal{Y}$ has a CP-rank at most $R$. Once $R$ is relatively small, $\mathcal{Y}$ is low-CP-rank.

Therefore, reconstructing $\mathcal{Y}$ from its partial entries can be formulated as a low-CP-rank tensor completion problem. There exist several methods available in the literature to solve this problem and generally they can be categorized into non-convex and convex methods. The weighted CP (WCP) decomposition method \cite{sue33,sue35} is a typical non-convex method, and it recovers tensors by solving the following non-convex optimization
\begin{equation}\label{eq:WCP}
\min\limits_{{{\bf{U}}^{\left( 1 \right)}}, \cdots, {{\bf{U}}^{\left( N \right)}}} \bigl\| {\mathcal{P}_{\Omega}\left( {{\cal Y} -  \bigl[\kern-0.25em\bigl[ {{\bf{U}}^{\left( 1 \right)}},{{\bf{U}}^{\left( 2 \right)}}, \cdots ,{{\bf{U}}^{\left( N \right)}} \bigr]\kern-0.25em\bigr] } \right)} \bigr\|_F^2,
\end{equation}
where ${{\bf{U}}^{\left( n \right)}} \in {\mathbb{C}^{{I_n} \times \hat R}}$ are the factor matrices, in which $\hat{R}$ is an estimated rank. Regarding convex methods, the following optimization was proposed \cite{sue41}
\begin{equation}\label{eq:tensornucmin}
\min_{\mathcal{X}}\|\mathcal{X}\|_*\quad\mbox{s.t.}\quad
\mathcal{P}_{\Omega}\mathcal{X}=\mathcal{P}_{\Omega}\mathcal{Y},
\end{equation}
where $\|\mathcal{X}\|_*$ is the tensor nuclear norm defined by
$$
\|\mathcal{X}\|_*=\sup_{\|\mathbf{a}_n\|_2=1,~1\leq n\leq N}\langle\mathcal{X},\mathbf{a}_1\circ\mathbf{a}_2\circ\ldots\circ\mathbf{a}_N\rangle.
$$
It was shown that the solution \eqref{eq:tensornucmin} will be $\mathcal{Y}$ if the number of known entries exceeds a certain amount \cite{sue41}. Unfortunately, \eqref{eq:tensornucmin} is NP-hard and computationally intractable. Recently some computational methods \cite{TNN, ADMM_R} are proposed to approximate or equally achieve the tensor nuclear norm.

% ---------------------- Low n-rank tensor completion ----------------
\subsection{Low-n-rank tensor completion}
The signal $\mathcal{Y}$ enjoys low $n$-rank as long as $R$ is small compared with $\min\{I_1,I_2,\ldots,I_N\}$, since the rank of the matricization $\mathbf{Y}_{(n)}$ of $\mathcal{Y}$ is at most $R$ if $R\leq I_n$. Thus, recovering $\mathcal{Y}$ from its partial entries can be viewed as a low-$n$-rank tensor completion problem. One finds a tensor that has a small $n$-rank \cite{sue31, sue34}
\begin{equation}\label{eq:lownrankcompletion}
\min\limits_{{\cal X}} \sum\limits_{n = 1}^N {{\alpha_n}{\rm{rank}}\left( {{{\bf{X}}_{\left( n \right)}}} \right)}, \ {\rm{s}}{\rm{.t}}{\rm{.}} \ {{\cal P}_\Omega }\left( {\cal X} \right) = {{\cal P}_\Omega }\left( {\cal Y} \right),
\end{equation}
where ${\alpha _n}$  is the weight and ${{\bf{X}}_{\left( n \right)}}$ denotes the mode-\emph{n} matricization.

One may approximate the non-convex objective in \eqref{eq:lownrankcompletion} by a convex one. The only non-convexity in \eqref{eq:lownrankcompletion} is the rank function. It is well known \cite{sue23,sue42} that the best convex approximation of the rank function is the nuclear norm, i.e., the sum of singular values. Also, to handle data with noise, the linear constraint \eqref{eq:lownrankcompletion} may be replaced by a least square fitting term. Altogether, one gets a convex optimization for low-$n$-rank tensor completion as follows
\begin{equation}\label{eq:nuclownrank}
\min\limits_{{\cal X}} {\kern 1pt} \sum\limits_{n = 1}^N \alpha _n{{{\left\| {{{\bf{X}}_{\left( n \right)}}} \right\|}_*}}  + \frac{\lambda}{2}\bigl\| {{{\cal P}_\Omega }\left( {\cal Y} \right) - {{\cal P}_\Omega }\left({{\cal X}} \right)} \bigr\|_F^2,
\end{equation}
where ${\left\| {{\kern 1pt} \cdot {\kern 1pt} } \right\|_*}$ denotes the nuclear norm of a matrix and $\lambda$ is the regularization parameter. When $\alpha_n=1$ for all $n=1,\ldots,N$, \eqref{eq:nuclownrank} is the convex model proposed in \cite{sue34}. Low-$n$-rank tensor completion has been successfully applied in computer vision \cite{sue31} and remote sensing data analyses \cite{sue34}.

%=====================================================
%----------------- The proposed method ---------------
%=====================================================
\section{The proposed method}\label{sec:HMRTC}

Though generic low-rank tensor completion methods discussed in Section \ref{sec:tensorcomp} are applicable to our $N$-D exponential signal reconstruction, they ignore the specific exponential structure of the factor vectors. From \eqref{eq:CPY}, it is observed that each factor matrix $\mathbf{A}^{\left( n \right)}$ defined in \eqref{eq:factormatrix} is \emph{Vandermonde matrix} and each factor vector in \eqref{eq:factor} is an exponential function. As a consequence, they will need unnecessarily large number of measurements for a stable reconstruction of $\mathcal{Y}$. Take $N=3$ and $I_1=I_2=I_3=I$ as an example. As discussed in \cite{sue41}, \eqref{eq:tensornucmin} needs $O\left(R^{1/2}(I\log I)^{3/2}\right)$ known entries to give a robust recovery for low-CP-rank tensor completion, and matrix completion theory suggests that $O(RI^2\log^2(I))$ observed entries are necessary for a reliable recovery for low-$n$-rank tensor completion \eqref{eq:nuclownrank}. However, there are only $O(R)$ degree of freedoms in the signal model \eqref{eq:signalmodel}. Therefore, it is expected that, if we explore the exponential structure of the factor vectors, we can design an $N$-D exponential signal reconstruction method that requires much fewer measurements than generic low-rank tensor completion methods.

In the following, we propose an approach that utilizes the exponential structure of the factor vectors, in addition to the low-CP-rank structure of the signal.

% ------------------- The proposed model --------------------
\subsection{The proposed model}
We will use the low-CP-rank structure of the tensor $\mathcal{Y}$, although the low-$n$-rank property is applicable as well. To promote the exponential structure of the factor vector, we enforce the Hankel matrix of each factor vector to be low-rank by nuclear norm. We propose the following reconstruction model,
\begin{equation}\label{eq:proposed}
\begin{split}
&\min\limits_{\substack{{\bf{u}}_r^{(n)}\\ r=1,\ldots,\hat{R} \\ n=1,\ldots,N}} \sum\limits_{r= 1}^{\hat R} {\sum\limits_{n = 1}^N {{{\bigl\| {{\cal R}{\bf{u}}_r^{( n )}} \bigr\|}_*}} }\\
&+\frac{\lambda }{2}\bigl\| {{{\cal P}_\Omega }\bigl( {\sum\limits_{r = 1}^{\hat R} {{\bf{u}}_r^{( 1 )} \circ {\bf{u}}_r^{( 2 )} \circ  \cdots  \circ {\bf{u}}_r^{( N )}} } \bigr) - {{\cal P}_\Omega }\left( {\cal Y} \right)} \bigr\|_F^2,
\end{split}
\end{equation}
where $\lambda $ is a regularization parameter that trades off the nuclear norm against the data consistency and $\mathcal{R}$ is a linear operator defined as $\mathcal{R}: \mathbb{C}^{I_n}\to\mathbb{C}^{S_1^{(n)}\times S_2^{(n)}}$, for some integers $S_1^{(n)}$ and $S_2^{(n)}$ satisfying $S_1^{(n)}+S_2^{(n)}=I_n+1$, as follows
$$
[\mathcal{R}{\bf{a}}_r^{(n)}]_{k,l}=[{\bf{a}}_{r}^{(n)}]_{k+l-1},\quad\forall~~1\leq k\leq S_1^{(n)},~1\leq l\leq S_2^{(n)}.
$$
The $\mathit{S}_1^{(n)}$ is chosen to make the Hankel matrix square or approximately square to minimize the reconstruction error \cite{sue30}. We call the proposed model \eqref{eq:proposed} Hankel Matrix nuclear norm Regularized low-CP-rank Tensor Completion (HMRTC).

The proposed HMRTC simultaneously exploits the low-CP-rank structure and the exponential structure of the associated factor vectors, which makes it superior to existing methods. 1) HMRTC utilizes the exponential structure of factor vectors, being demonstrated in Section \ref{sec:exp} that HMRTC can significantly reduce the number of necessary samples, compared with generic low-rank tensor completion methods that do not consider the structure of the factor matrices. 2) HMRTC represents the unknown tensor in a CP decomposition form, significantly reducing the size of variables in numerical algorithms, compared with atomic norm minimization (e.g. \cite{sue18}) and other low-rank structured matrix methods (e.g. EMaC \cite{sue30}). In particular, as we will see in Section \ref{sec:Numerical algorithm}, the numerical algorithm of HMRTC invokes $O(\hat{R}N)$ matrices of size $O(I^2)$ if $I_1=\ldots=I_N=I$. As a comparison, the standard alternating direction method of multipliers (ADMM) algorithm for solving atomic norm minimization and EMaC needs $O(1)$ huge matrices of size $O(I^{2N})$. Consequently, our proposed HMRTC can easily reconstruct $3$-D exponential signals of size $50\times 50\times 50$, which, however, is a prohibitive task for EMaC or atomic norm minimization.
%3) The proposed method does not explicitly enforce the Hankel matrix of each factor vector to be rank-1. This is flexible %since low-CP-rank decomposition may produce one or multiple exponentials in the each factor vector\cite{MUNIN}.

\begin{figure*}[htbp]
\centering
\includegraphics[width=6in]{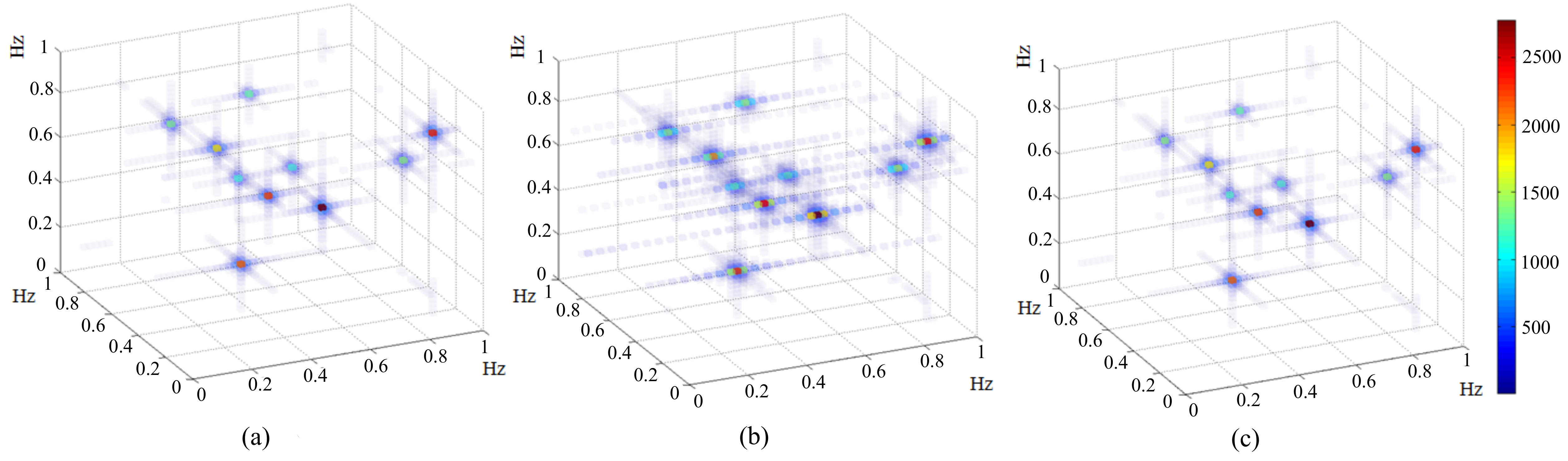}
\caption{The 3-D spectra recovered by the HMRTC from a tensor with half of slices missing. (a) The ground truth spectrum; (b) Reconstructed spectrum by filling zeroes into missing data points in the time domain and then performing Fourier transform on the full time domain signal; (c) Reconstructed spectrum by HMRTC. Note: We simulate a tensor with 10 damped complex sinusoids and discard half of slices. We draw $3$-D spectra of tensors by using the tensor toolbox \cite{sue58}. The color stands for magnitude of spectra.}
\label{RecSpec_MissingSlicesTensor}
\end{figure*}

We conclude this subsection by a toy example to demonstrate the great potential of the proposed HMRTC. It is well known that neither the low-$n$-rank tensor completion nor WCP can recover missing slices \cite{sue23, sue33} since the information is unknown in the missing part. On the contrary, HMRTC is able to estimate these missing data because the exponential structure of one single factor is exploited. Fig. \ref{RecSpec_MissingSlicesTensor} shows that HMRTC can perform well even when half of the slices are missing. Thus, HMRTC will be useful for those applications requiring recover the truncated data in the end of signals to improve the frequency resolution or increase the signal-to-noise ratio \cite{sue13, sue44}.

%\begin{figure}[htbp]
%\centering
%\includegraphics[width=3in]{Tensors_MissingSlices.eps}
%\caption{A superposition of rank-1 tensors to fit the tensor with missing slices (in black).}
%\label{Tensors_MissingSlices}
%\end{figure}

% ----------------- Numerical algorithm ------------------
\subsection{Numerical algorithm}\label{sec:Numerical algorithm}
By combining the unknown factors ${\bf{u}}_n^{(r)}$ into factor matrices
${{\bf{U}}^{(n)}} = \bigl[
{\bf{u}}_1^{( n )}\ldots {\bf{u}}_R^{(n)}
 \bigr]$ for $n=1,\ldots,N$,
Eq. \eqref{eq:proposed} is rewritten concisely as
\begin{equation}\label{eq:proposednew}
\begin{split}
\min\limits_{\substack{{{\bf{U}}^{( n )}}\\ n=1,\ldots,N} }&\sum\limits_{r = 1}^{\hat R} {\sum\limits_{n = 1}^N {{{\bigl\| {{\cal R}{{\cal Q}_r} {{\bf{U}}^{( n )}}} \bigr\|}_*}} }  \\
+ \frac{\lambda }{2}&\bigl\| {{{\cal P}_\Omega }( {\cal Y} ) - {{\cal P}_\Omega }\bigl( { \bigl[\kern-0.25em\bigl[ {{\bf{U}}^{( 1 )}},{{\bf{U}}^{( 2 )}}, \cdots ,{{\bf{U}}^{( N )}} \bigl]\kern-0.25em\bigl] } \bigr)} \bigr\|_F^2,
\end{split}
\end{equation}
where ${{\cal Q}_r}$ extracts the $r$-th column from ${{\bf{U}}^{\left( n \right)}}$ for $r=1,\ldots,R$ and $n = 1, \ldots ,N$.

Recently, it has been shown in \cite{sue32} and \cite{ADMM} that the ADMM is very efficient for some convex or nonconvex problems in various applications. To solve (\ref{eq:proposednew}), we also propose an algorithm based on ADMM. Some auxiliary variables, ${\bf{Z}}_r^{(n)} = {{\cal R}}{Q_r}{{\bf{U}}^{(n)}}$, $n = 1, \ldots ,N$ and $r = 1, \ldots ,\hat R$ are introduced and then (\ref{eq:proposednew}) is reformulated into the following equivalent form:
\begin{equation}\label{eq:ADMM_1}
\begin{aligned}
&\min\limits_{\substack{{{\bf{U}}^{( n )}}\\ n=1,\ldots,N\\ r=1,\ldots,{\hat R}}}\sum\limits_{r = 1}^{\hat R} {\sum\limits_{n = 1}^N {{{{\big\| {{\bf{Z}}_r^{(n)}} \big\|}_*}}}}\\
&{\kern 20pt}+\frac{\lambda }{2}\big\| {{{{\cal P}}_\Omega }({{\cal Y}}) - {{{\cal P}}_\Omega }(\bigl[\kern-0.25em\bigl[ {{{\bf{U}}^{(1)}},{{\bf{U}}^{(2)}}, \ldots ,{{\bf{U}}^{(N)}}} \bigr]\kern-0.25em\bigr])} \big\|_F^2\\
&{\kern 10pt}{\rm{s}}{\rm{.t}}{\rm{.}} \, {\bf{Z}}_r^{(n)} = {{\cal R}}{Q_r}{{\bf{U}}^{(n)}},\forall {\kern 3pt}1 \le r \le \hat R,{\kern 5pt} 1 \le n \le N.% {\rm{for}} \: r = 1{\rm{,}} \ldots {\rm{,}}\hat R \, {\rm{and}} \, n = 1, \ldots ,N.
\end{aligned}
\end{equation}

For ease of presentation, we define
$${{\cal U}} = \big\{ {{{\bf{U}}^{(1)}}, \ldots ,{{\bf{U}}^{(N)}}} \big\},$$
$${{\cal Z}} = \big\{ {{\bf{Z}}_1^{(1)}, \ldots ,{\bf{Z}}_1^{(N)}, \ldots ,{\bf{Z}}_{\hat R}^{(1)}, \ldots ,{\bf{Z}}_{\hat R}^{(N)}} \big\},$$
and
$${{\cal D}} = \big\{ {{\bf{D}}_1^{(1)}, \ldots ,{\bf{D}}_1^{(N)}, \ldots ,{\bf{D}}_{\hat R}^{(1)}, \ldots ,{\bf{D}}_{\hat R}^{(N)}} \big\}.$$

The augmented Lagrangian function of \eqref{eq:ADMM_1} is
\begin{equation}\label{eq:ADMM_2}
\begin{aligned}
& {{{\cal L}}_\beta }({{\cal U}}, {{\cal Z}}, {{\cal D}}) = \sum\limits_{r = 1}^{\hat R} {\sum\limits_{n = 1}^N {\big ({{\big\langle {{\bf{D}}_r^{(n)},{{\cal R}}{Q_r}{{\bf{U}}^{(n)}} - {\bf{Z}}_r^{(n)}} \big\rangle }} }} \\
& {\kern 10pt}+ {{\bigl\| {{\bf{Z}}_r^{(n)}} \bigr\|}_*} + {{\frac{\beta }{2}\bigl\| {{{\cal R}}{Q_r}{{\bf{U}}^{(n)}} - {\bf{Z}}_r^{(n)}} \bigr\|_F^2 \big )} } \\
& {\kern 10pt}+\frac{\lambda }{2}\bigl\| {{{{\cal P}}_\Omega }({{\cal Y}}) - {{{\cal P}}_\Omega }(\bigl[\kern-0.25em\bigl[ {{{\bf{U}}^{(1)}},{{\bf{U}}^{(2)}}, \ldots ,{{\bf{U}}^{(N)}}} \bigr]\kern-0.25em\bigr])} \bigr\|_F^2,
\end{aligned}
\end{equation}
where ${\bf{D}}_r^{(n)}$ is the matrix of Lagrange multipliers for $n = 1, \ldots ,N$ and $r = 1, \ldots ,\hat R$.

The ADMM is an iterative algorithm. Given ${{\cal U}}_k$, ${{\cal Z}}_k$, and ${{\cal D}}_k$ at step $k$, it updates ${{\cal U}}$, ${{\cal Z}}$, and ${{\cal D}}$ as follows.

\paragraph{}
The variable ${{\cal U}}$ is updated by solving the following optimization,
\begin{equation}\label{eq:ADMM_3}
\mathop {\min }\limits_{{\cal U}} {{{\cal L}}_{{\beta _k}}}({{\cal U}},{{{\cal Z}}_k}, {{{\cal D}}_k}).
\end{equation}
%where ${{{\cal Z}}_k}$ and ${{{\cal D}}_k}$ are ${{\cal Z}}$ and ${{\cal D}}$ respectively at the $k$.
Due to the multi-linearity of the CP decomposition, it is not easy to solve \eqref{eq:ADMM_3} exactly. Here, we employ an alternating minimization procedure to solve \eqref{eq:ADMM_3} approximately.
Fixing ${{\bf{U}}^{(1)}},\ldots,{{\bf{U}}^{(n-1)}}, {{\bf{U}}^{(n+1)}},\ldots,{{\bf{U}}^{(N)}}$, we solve \eqref{eq:ADMM_3} with respect to ${{\bf{U}}^{(n)}}$, which is a convex optimization as follows:
\begin{equation}\label{eq:ADMM_4}
\begin{aligned}
& \min\limits_{{{\bf{U}}^{(n)}}} \sum\limits_{r = 1}^{\hat R} {(\frac{{{\beta_k}}}{2}\bigl\| {{{\cal R}}{Q_r}{{\bf{U}}^{(n)}} - {\bf{Z}}_{r;k}^{(n)} + {{({\beta_k})}^{ - 1}}{\bf{D}}_{r;k}^{(n)}} \bigr\|_F^2)}  \\
& + \frac{\lambda }{2}\bigl\| {{{{\cal P}}_{{\Omega ^{(n)}}}}({{\bf{Y}}_{(n)}}) - {{{\cal P}}_{{\Omega ^{(n)}}}}({{\bf{U}}^{(n)}}{\bf{G}}_k^{\left( n \right)})} \bigr\|_F^2,
\end{aligned}
\end{equation}
where ${\bf{G}}_k^{\left( n \right)} = ({\bf{U}}_k^{\left( N \right)} \odot  \cdots  \odot {\bf{U}}_k^{\left( {n + 1} \right)} \odot {\bf{U}}_{k + 1}^{\left( {n - 1} \right)} \odot  \cdots  \odot {\bf{U}}_{k + 1}^{\left( 1 \right)})^T$; ${\bf{D}}_{r;k}^{(n)}$ and ${\bf{Z}}_{r;k}^{(n)}$ are the \textit{k}-th update of ${\bf{D}}_r^{(n)}$ and ${\bf{Z}}_r^{(n)}$;
${{\bf{Y}}_{(n)}}$ and ${{{\cal P}}_{{\Omega ^{(n)}}}}({{\bf{Y}}_{(n)}})$ are the mode-$n$ matricization of the tensors ${{\cal Y}}$ and ${{{\cal P}}_\Omega }({{\cal Y}})$ respectively.
It is obvious \eqref{eq:ADMM_4} is a least squares problem, and therefore its solution is a solution of linear system. In particular, if we define ${{\cal A}}_k^{\left( n \right)}\left( {\bf{X}} \right) \buildrel \Delta \over = {{{\cal P}}_{{\Omega ^{\left( n \right)}}}}\left( {{\bf{XG}}_k^{\left( n \right)}} \right)$,
then \eqref{eq:ADMM_4} is rewritten as
\begin{equation}\label{eq:ADMM_10}
\begin{aligned}
&\min\limits_{{{\bf{U}}^{(n)}}} \sum\limits_{r = 1}^{\hat R} {(\frac{{{\beta _k}}}{2}\bigl\| {{\cal R}{Q_r}{{\bf{U}}^{(n)}} - {\bf{Z}}_{r;k}^{(n)} + {{({\mu ^k})}^{ - 1}}{\bf{D}}_{r;k}^{(n)}} \bigr\|_F^2)}\\
&{\kern 20pt}+ \frac{\lambda }{2}\bigl\| {{{\cal P}_{{\Omega ^{(n)}}}}({{\bf{Y}}_{(n)}}) - {\cal A}_k^{\left( n \right)}\left( {{{\bf{U}}^{(n)}}} \right)} \bigr\|_F^2,
\end{aligned}
\end{equation}

whose solution satisfies
\begin{equation}\label{eq:ADMM_5}
\begin{aligned}
& \lambda {{\cal A}}_k^{\left( n \right)*}{{\cal A}}_k^{\left( n \right)}{{\bf{U}}^{\left( n \right)}} + \beta_k \sum\limits_{r = 1}^{\hat R} {{{{\cal Q}}_r}{{\cal R}{\cal R}}{{{\cal Q}}_r}{{\bf{U}}^{\left( n \right)}}} \\
&{\kern 20pt}= \lambda {{{\cal A}}_k^{\left( n \right)*}}\left( {{{{\cal P}}_{{\Omega ^{(n)}}}}({{\bf{Y}}_{(n)}})} \right)  \\
&{\kern 30pt}+\beta_k \sum\limits_{r = 1}^{\hat R} {{{\cal Q}}_r^*{{{\cal R}}^*}\left( {{\bf{Z}}_{r;k}^{(n)} - {{({\beta_k})}^{ - 1}}{\bf{D}}_{r;k}^{(n)}} \right)},
\end{aligned}
\end{equation}

The closed-form of ${{\bf{U}}^{(n)}}$ are derived in Appendix A.

\paragraph{}
The variable ${{\cal Z}}$ is updated via the following optimization
\begin{equation}\label{eq:ADMM_6}
\min\limits_{{\cal Z}} {{{\cal L}}_{{\beta _k}}}({{{\cal U}}_{k+1}},{{\cal Z}},{{{\cal D}}_k}).
\end{equation}
which is equivalent to $N\hat{R}$ independent sub-problems
\begin{equation}\label{eq:ADMM_7}
\min\limits_{{{\bf{Z}}^{\left( {n} \right)}_r}} {\bigl\| {{\bf{Z}}_r^{(n)}} \bigr\|_*} + \frac{{{\beta _k}}}{2}\bigl\| {{{\cal R}}{Q_r}{\bf{U}}_{k + 1}^{\left( n \right)} - {\bf{Z}}_r^{(n)} + {{({\beta _k})}^{ - 1}}{\bf{D}}_{r;k}^{(n)}} \bigr\|_F^2.
\end{equation}
for $n = 1, \ldots ,N$ and $r = 1, \ldots ,\hat R$.
Following \cite{sue42}, the closed-form solution of \eqref{eq:ADMM_7} is
\begin{equation}\label{eq:ADMM_8}
\mathbf{Z}_r^{\left( n \right)} = {S_{1/{\beta_k}}}\left( {{{\cal R}}{{{\cal Q}}_r}{\bf{U}}_{k + 1}^{\left( n \right)} + {{({\beta _k})}^{ - 1}}{\bf{D}}_{r;k}^{(n)}} \right),
\end{equation}
where $S_{1/{\beta_k}}$ is the soft singular value thresholding operator with parameter $1/\beta_k $.

\paragraph{}
Update ${\bf{D}}_r^{\left( n \right)}$ by
\begin{equation}\label{eq:ADMM_9}
{\bf{D}}_{r;k + 1}^{\left( n \right)} = {\bf{D}}_{r;k}^{\left( n \right)} + {\beta _k}\left( {{{\cal R}}{Q_r}{\bf{U}}_{k + 1}^{\left( n \right)} - {\bf{Z}}_{r;k + 1}^{(n)}} \right).
\end{equation}

The full algorithm is described in Algorithm \ref{alg:1}. This algorithm can also be accelerated by setting ${\beta _{k + 1}} = \rho {\beta _k}$, where $\rho  \in (1.0,\, 1.1]$ \cite{sue31, sue32, ma}.

\begin{algorithm}[!ht]
\caption{The pseudo code of the proposed algorithm}\label{alg:1}
\begin{algorithmic}[1]
\REQUIRE  The tensor ${{\cal Y}} \in {\mathbb{C}^{{I_1} \times  \cdots  \times {I_N}}}$, the set $\Omega$, and parameters $\hat R$, $\lambda$;
\ENSURE The reconstruction ${\cal X} =  \bigl[\kern-0.25em\bigl[ {{\bf{U}}^{( 1 )}},{{\bf{U}}^{( 2 )}}, \cdots ,{{\bf{U}}^{( N )}} \bigr]\kern-0.25em\bigr]$ of $\mathcal{Y}$;
\STATE Initialize ${\bf{U}}^{\left( 1 \right)}, \ldots ,{\bf{U}}^{\left( N \right)}$, $k = \Delta x = 1$, ${\beta _0} = 0.1$ and $\rho  = 1.05$.
\WHILE {$\Delta x < {10^{ - 4}}$ and $k < {10^3}$}
\STATE Update ${{\bf{U}}^{\left( 1 \right)}},{{\bf{U}}^{\left( 2 \right)}}, \cdots ,{{\bf{U}}^{\left( N \right)}}$ by solving \eqref{eq:ADMM_5};
\STATE Update ${\bf{Z}}_r^{(n)}$ by solving \eqref{eq:ADMM_8} for $r = 1, \ldots ,\hat R$, $n = 1, \ldots ,N$;
\STATE Update ${\bf{D}}_r^{(n)}$ by solving \eqref{eq:ADMM_9} for $r = 1, \ldots ,\hat R$, $n = 1, \ldots ,N$;
\STATE Update $\beta $ by ${\beta _{k{\rm{ + }}1}}{\rm{ = }}\rho {\beta _k}$;
\STATE ${{\cal X}} = \bigl[\kern-0.25em\bigl[ {{{\bf{U}}^{\left( 1 \right)}},{{\bf{U}}^{\left( 2 \right)}}, \cdots ,{{\bf{U}}^{\left( N \right)}}} \bigr]\kern-0.25em\bigr]$;
\STATE $\Delta x = {\left\| {{{\cal X}} - {{{\cal X}}_{{\rm{last}}}}} \right\|_F}/{\left\| {{{{\cal X}}_{{\rm{last}}}}} \right\|_F}$; ${{{\cal X}}_{{\rm{last}}}} \leftarrow {{\cal X}}$;
\STATE $k = k + 1$;
\ENDWHILE
\end{algorithmic}
\end{algorithm}

% ----------------- Convergence analysis ------------------
\subsection{Convergence analysis}
We provide the convergence of Algorithm \ref{alg:1} stated in \textbf{Theorem 1} and \textbf{Theorem 2}.
From the theorems, we know that the sequence $\{\mathcal{U}_k\}$ generated by Algorithm \ref{alg:1} converges. Furthermore, if we further impose some condition on the Lagrange multipliers $\{\mathcal{D}_k\}$, then the limit is a critical point of \eqref{eq:proposednew}.

\begin{theorem}
The sequence $\{\mathcal{U}_k\}$ generated by Algorithm \ref{alg:1} is a Cauchy sequence.
\end{theorem}

\begin{theorem}
If $\mathop {\lim }\limits_{k \to \infty } \bigl\| {{\bf{D}}_{r;k + 1}^{\left( n \right)} - {\bf{D}}_{r;k}^{\left( n \right)}} \bigr\|_F = 0$, for all $r \in \{ 1, \ldots ,\hat R\}$, $n \in \{ 1, \ldots ,N\}$, then the limit of $\{\mathcal{U}_k\}$ satisfies the KKT condition for \eqref{eq:proposednew}.
\end{theorem}

Proofs of \textbf{Theorem 1} and \textbf{Theorem 2} are added in Appendix C.

% ----------------- Complexity analysis ------------------
\subsection{Complexity analysis}
The computational complexity of HMRTC is analyzed here. Besides the typical tensor operation such as Tucker operation, the running time of Algorithm \ref{alg:1} is dominated by the singular value decomposition (SVD) for the singular value thresholding operator in \eqref{eq:ADMM_8}. Consider to recover a tensor ${{\cal X}} \in {\mathbb{C}^{{I_{\rm{1}}} \times  \cdots  \times {I_N}}}$ with ${I_{\rm{1}}}{\rm{ = }} \ldots {\rm{ = }}{I_N}{\rm{ = }}I$. The SVD of ${\bf{Z}}_r^{\left( n \right)}$, which is of a small size $0.5I \times 0.5I$, can be done in $O(I^3)$ operations. Since we have $N\hat R$ SVDs to compute at each iteration, the total computational complexity for SVD in each iteration is $O\big( {N\hat RI^3} \big)$, which is only sub-linear with the tensor size ${I^N}$ when $N \ge 3$. Furthermore, the SVDs in a single iteration can be computed in parallel, since each ${\bf{Z}}_r^{\left( n \right)}$, $r = 1, \ldots ,\hat R$, $n = 1, \ldots ,N$, is updated independently. Therefore, HMRTC has the potential to be applied in $N$-D $\left( {N \ge 3} \right)$  exponential signal recovery shown in Section \ref{sec:high-dimension}. In addition, from the spatial complexity point of view, the storage $O\big( {N\hat RI} \big)$ of HMRTC can be significantly smaller than that of the original tensor when the estimated tensor rank $\hat R$ is much smaller than $I$.

%===========================================================
%-------------------- EXPERIMENTAL RESULTS -----------------
%===========================================================
\section{EXPERIMENTAL RESULTS}\label{sec:exp}
In this section, we will evaluate the proposed HMRTC on simulated exponential signals, including undamped and damped complex sinusoids, and real NMR spectroscopy data. Two state-of-the-art algorithms of low-rank tensor completion, the alternating direction method-based tensor recovery (ADM-TR) \cite{sue34} and WCP \cite{sue33}, are compared with HMRTC.

The parameters of HMRTC are listed in Algorithm \ref{alg:1}. For ADM-TR and WCP, the maximum number of iterations is ${10^3}$. In ADM-TR, the parameters are ${c_\lambda } = {c_\beta } = 1$ and $\beta = 1$. The alternating least squares method is applied to solve WCP and the algorithm is terminated when ${{\rm{abs}}\left( {{f_k} - {f_{k + 1}}} \right)}/\big(1 + {f_k}\big) \le {10^{ - 6}}$, where ${f_k}$ is the objective function value in \eqref{eq:WCP} after iteration $k$.

% ---------------- Experiments setup ------------------
\subsection{Experiment setup for simulated data}
The proposed algorithm is tested on $3$-D simulated signals ${\cal Y} \in {\mathbb{C}^{50 \times 50 \times 50}}$ as follows. The frequency ${f_{n, r}}$ is uniform randomly drawn from the interval $[0,1)$  for all $1\leq r\leq R$ and $1\leq n\leq 3$, where $R$ is the number of exponentials. The coefficient ${d_r}$  is generated by ${d_r} = 1 + {10^{0.5{m_r}}}$ and the damping factor ${\tau _{n, r}}$ is generated by ${\tau _{n,r}} = 10 + 30{g_{n,r}}$, where ${m_r}$ and ${g_{n,r}}$ follow the uniformly random distribution on $[0,1]$. The undamped complex sinusoid is synthesized as
\begin{equation}\label{eq:signalgenerator}
{y_{{i_1},{i_2},{i_3}}} = \sum\limits_{r = 1}^R {{d_r}{{\left( {{e^{j2\pi {f_{1,r}}}}} \right)}^{{i_1}}}{{\left( {{e^{j2\pi {f_{2,r}}}}} \right)}^{{i_2}}}{{\left( {{e^{j2\pi {f_{3,r}}}}} \right)}^{{i_3}}}},
\end{equation}
and damped complex sinusoid is simulated as
\begin{equation}\label{eq:signalgenerator_damp}
\begin{split}
{y_{{i_1},{i_2},{i_3}}} &= \sum\limits_{r = 1}^R {{d_r}{{\left( {{e^{ - 1/{\tau _{1,r}} + 2\pi j{f_{1,r}}}}} \right)}^{{i_1}}}}\\
&{\kern 20pt}{{\left( {{e^{ - 1/{\tau _{2,r}} + 2\pi j{f_{2,r}}}}} \right)}^{{i_2}}}
{{\left( {{e^{ - 1/{\tau _{3,r}} + 2\pi j{f_{3,r}}}}} \right)}^{{i_3}}},
\end{split}
\end{equation}
for ${i_n} = 0, \cdots ,49$ and $n $=1, 2, 3.

In each experiment, the simulated ground-truth tensor ${\cal Y}$  is normalized by dividing the maximum magnitude of its entries. Gaussian white noise with standard deviation $\sigma$ is added on both real and imaginary parts of $\mathcal{Y}$.

We denote by $\cal X$ the reconstruction of HMRTC, i.e., the output of Algorithm \ref{alg:1}, and the relative least normalized error (RLNE) is defined as
\begin{equation}
{\rm{RLNE}} = {\| {{\cal X} - {\cal Y}} \|_F}/{\| {\cal Y} \|_F}.
\end{equation}
The average RLNE is calculated by averaging the RLNEs over $10$ Monte Carlo trials as conducted in \cite{sue31, sue52}. To facilitate comparison in color map, the average RLNE is set to be $1$ if it is larger than $1$. In each trial, the observed entries are sampled in the uniformly random fashion and the sampling ratio ($SR$) is denoted as the proportion of available data to the full data.

The numerical experiments are conducted on a Dell PC running Windows 7 operating system with Intel Core i7 2600 CPU and $12$-GB RAM. The average computation time for ADM-TR, WCP and HMRTC to recover simulated signals with $R = 50$ and $SR = 0.5$ is $185$s, $7395$s and $675$s, respectively. The average memory requirements for ADM-TR, WCP and HMRTC is $0.69$GB, $0.65$GB and $0.66$GB, respectively.

% ---------------------- Robustness to the estimated tensor rank --------------------
\subsection{Robustness to the estimated tensor rank}
We first evaluate the robustness of WCP and HMRTC to the estimated tensor rank. ADM-TR is not compared since it does not require an estimation of the tensor rank.

\begin{figure}[htb]
\centering
\includegraphics[width=3.4in]{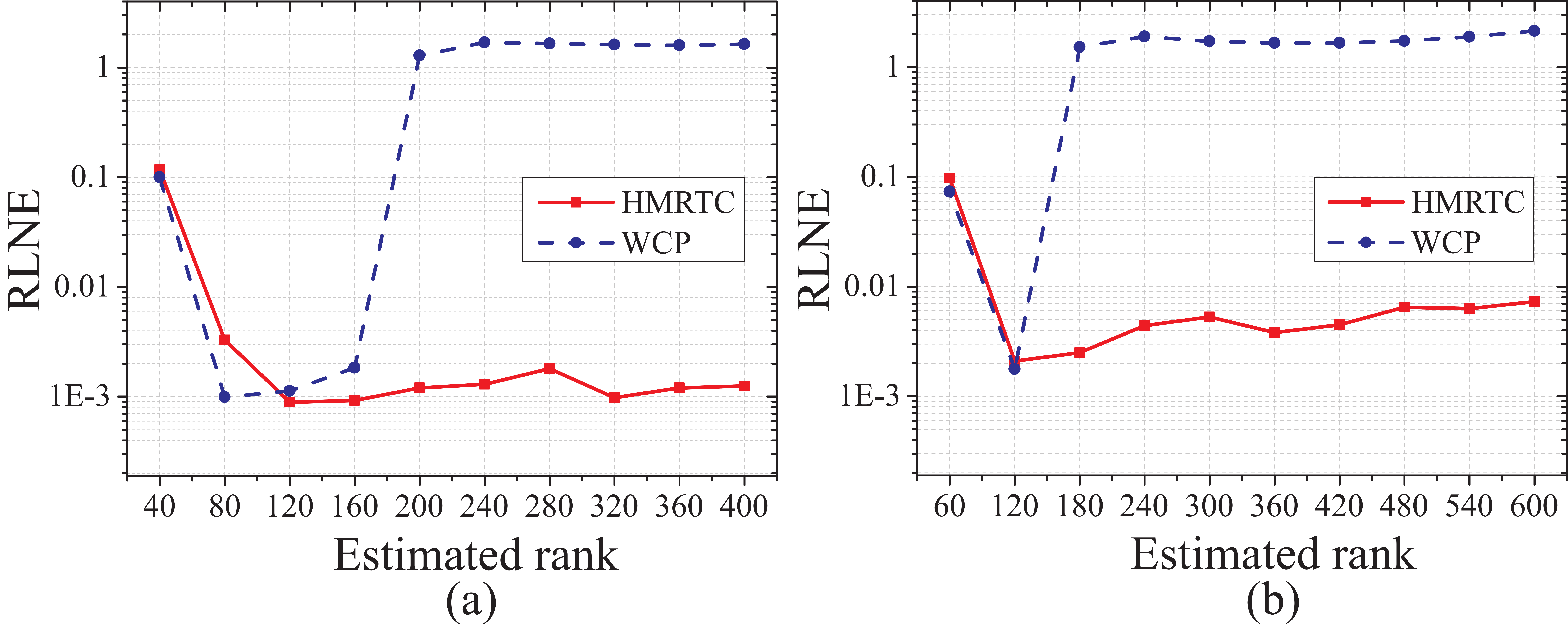}
\caption{RLNE versus the estimated tensor rank. (a) and (b) illustrate reconstruction errors of WCP and HMRTC with different estimated ranks $\hat R$ chosen from $\left\{ {R,2R, \ldots ,10R} \right\}$. The sampling ratio $SR$ is $0.3$ and the number of exponentials $R$ in (a) and (b) is $40$ and $60$, respectively. Here $\rho=1.02$.}
\label{RLNE_vs_R_hat}
\end{figure}

The comparison is presented in Fig. \ref{RLNE_vs_R_hat}. It is observed that the HMRTC can always achieve low reconstruction errors (RLNE$\leq$0.1) as $\hat R$ increases from $2R$ to $10R$, while WCP fails to recover if $R$ is over-estimated too much. In particular, when the ground truth $R = 60$, as shown in Fig. \ref{RLNE_vs_R_hat}(b), WCP can work only when $\hat R = 2R$ but HMRTC is robust to any $\hat R$ chosen from $2R$ to $10R$. Hence, HMRTC will be more useful for those applications where it is difficult to estimate the true number of exponentials. In the following simulated experiments, we set $\hat R=2R$ in WCP and $\hat{R}=100$ in HMRTC.

% ----------------------- Recovery of simulated undamped complex sinusoids -----------------
\subsection{Recovery of simulated complex sinusoids}\label{sec:simulation}
In this subsection, undamped and damped complex sinusoids are simulated to evaluate the construction performance. The regularization parameter $\lambda $ in ADM-TR is set be ${10}$, and be ${10^2}$ and ${10^3}$ in HMRTC for data with noise level $0.01$ and $0.005$, respectively.

\begin{figure*}[htb]
\centering
\includegraphics[width=5.5in]{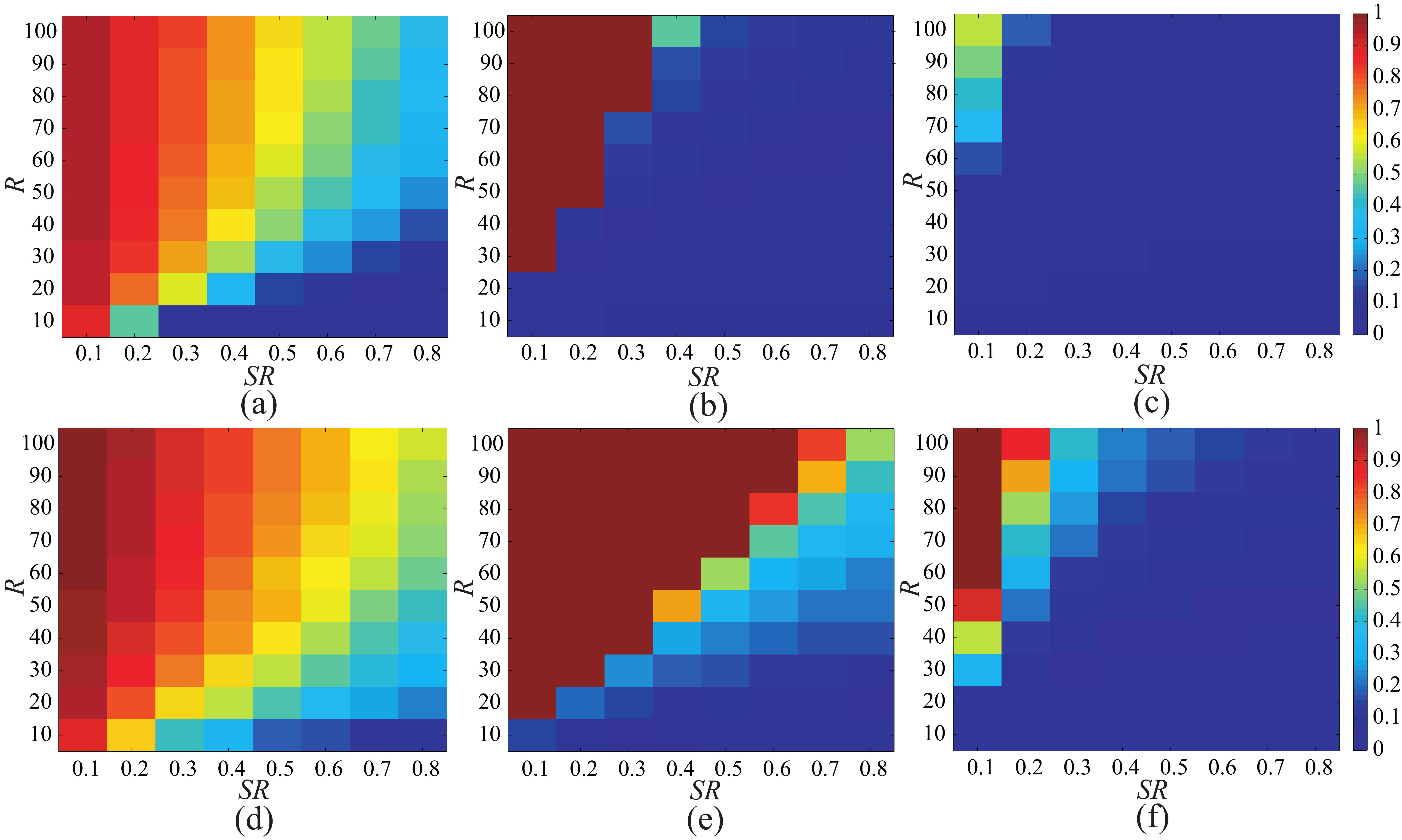}
\caption{Reconstruction errors of ADM-TR, WCP and HMRTC. (a), (b) and (c) reflect the average RLNEs by ADM-TR, WCP and HMRTC, respectively, in recovering undamped complex sinusoids with noise level $\sigma  = 0.01$. (d), (e) and (f) indicate the average RLNEs by ADM-TR, WCP and HMRTC, respectively, in recovering damped complex sinusoids with noise level $\sigma  = 0.005$.}
\label{Phase_Transition}
\end{figure*}

Fig. \ref{Phase_Transition} shows that HMRTC yields an average RLNE that is much smaller than ADM-TR and WCP for all $R$'s and $SR$'s, no matter in recovering undamped or damped sinusoids. Furthermore, for a fixed $R$, HMRTC needs a much smaller $SR$ than ADM-TR and WCP to achieve an average RLNE within noise level, implying that HMRTC requires a much smaller number of sampled entries for the stable recovery than ADM-TR and WCP.
%This result is not surprising: Although WCP recovers tensors by utilizing the low-CP-rank structure while ADM-TR completes the %tensor under the low-$n$-rank assumption, both of them ignore the exponential structure of the associated factor vectors.

\begin{figure}[ht]
\centering
\includegraphics[width=3.4in]{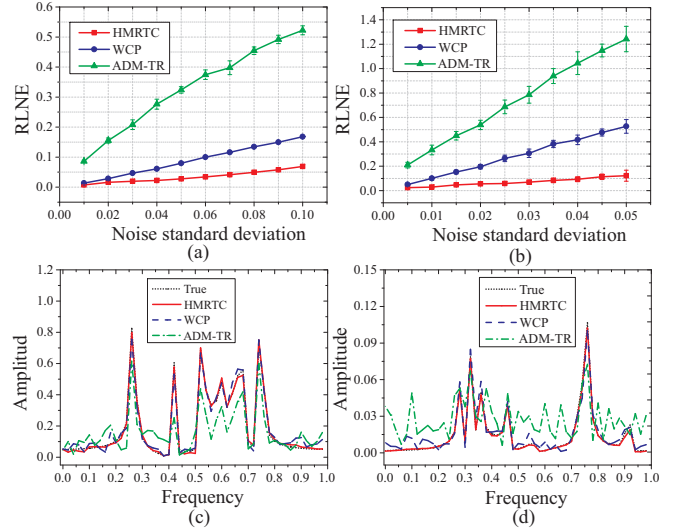}
\caption{Reconstruction results of ADM-TR, WCP and HMRTC from noisy data: (a) and (b) The average RLNE versus noise standard deviation $\sigma$ in recovering undamped and damped complex sinusoid with $R = 10$, respectively. The sampling ratio $SR$ is set to be 0.3 and 0.5 in (a) and (b), respectively. (c) and (d) The spectra of chosen fibers in reconstruction in the undamped case with $\sigma  = 0.1$ and damped case with $\sigma  = 0.025$, respectively.}
\label{noise_level}
\end{figure}

Fig. \ref{noise_level} further examines the stability of the proposed algorithm at different noise levels. It is shown that HMRTC produces reconstructions with the lowest RLNE, as well as  the smallest variance, among the three methods, implying that HMRTC is the most robust to noise. The spectra of selected fibers in the reconstruction in Fig. \ref{noise_level}(c) and Fig. \ref{noise_level}(d) indicate that HMRTC leads to most consistent spectra to the ground-truth.

% --------------- Recovery of realistic NMR spectroscopy data ------------------
\subsection{Recovery of real NMR spectroscopy data}\label{sec:real}
NMR spectroscopy has been an indispensable tool in the study of structure, dynamics, and interactions of biopolymers in chemistry and biology. The duration of an \emph{N}-D NMR spectroscopy experiment is proportional to the number of measured data points, and the nonuniform sampling of time domain signal can dramatically reduce measurement time \cite{sue13, sue53, sue54, sue55}.Here we apply the proposed HMRTC to recover full spectrum in fast NMR, since the time domain signal of NMR is generally modeled as damped complex sinusoids\cite{sue13}.

\begin{figure*}[!htb]
\centering
\includegraphics[width=7in]{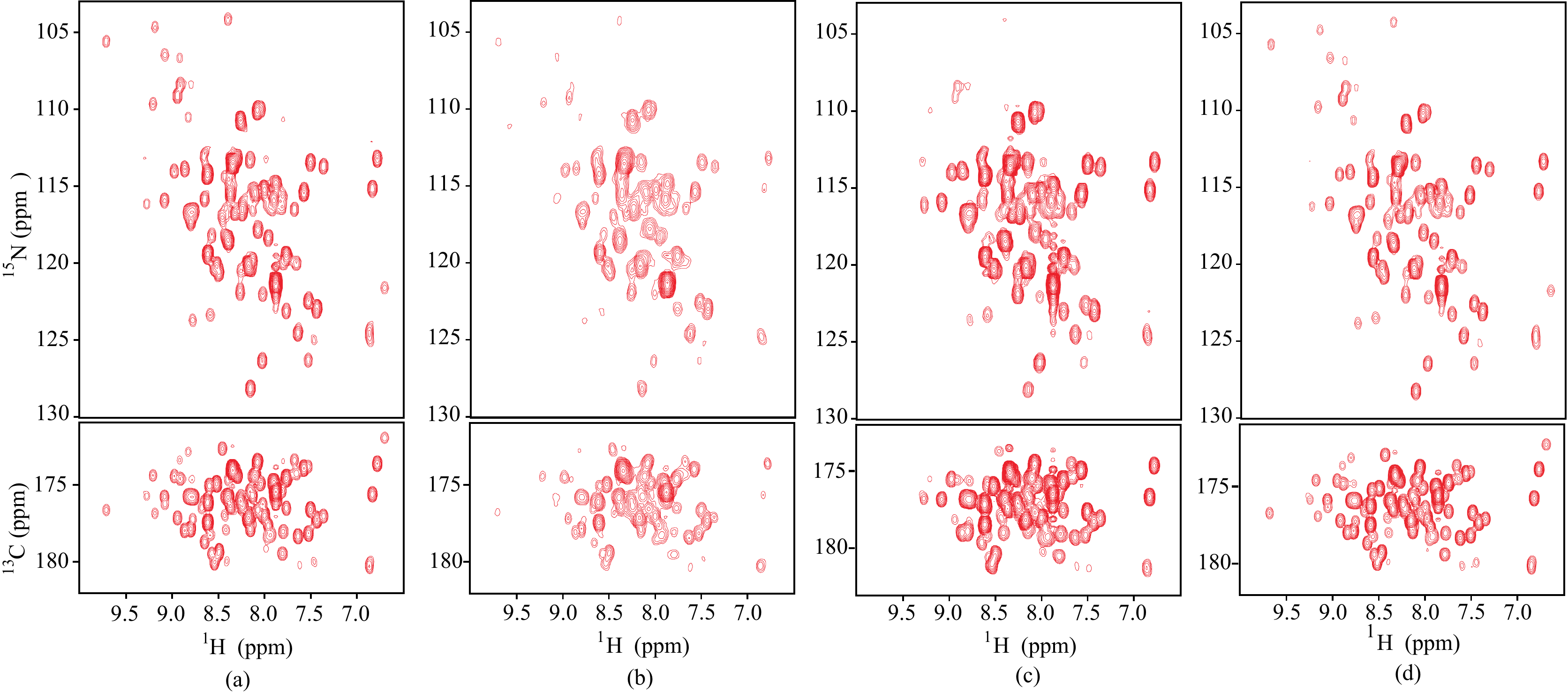}
\caption{The ${^1}$H-${^{15}}$N and ${^1}$H-${^{13}}$C skyline projection spectra in the 3-D HNCO experiment. (a) the uniformly-sampled spectrum; (b), (c) and (d) are the ADM-TR, WCP and HMRTC reconstruction using 10\% sampled data, respectively. The skyline projection spectra \cite{sue56}, plotted by summing all the frequency points along the rest dimensions, are used here to simplify the description of spectra. The experiment is carried out on a Bruker AVANCE III 600 MHz spectrometer equipped with a cryogenic probe at 293K. The estimated rank in WCP and HMRTC is $200$ and $500$, respectively. The $\lambda$ in ADM-TR and HRMTC is $10$ and ${10^3}$, respectively. The ppm denotes parts per million, the unit of chemical shift.}
\label{RealNMRSpectrum}
\end{figure*}

A $3$-D HNCO spectrum is tested and its sample is the U-[${^{15}}$N, ${^{13}}$C] RNA recognition motifs domain of protein RNA binding motif protein $5$ \cite{sue57}, which is a component of the spliceosome A-complex. The ADM-TR, WCP and HMRTC are compared in recovering this $3$-D spectrum with the size of 64$\times$128$\times$512 from a $3$-D Poisson-gap \cite{sue53} nonuniformly sampled time-domain data. All the spectra are processed in NMRPipe \cite{sue62} using a routine processing manner.

Fig. \ref{RealNMRSpectrum} shows that HMRTC leads to the most faithful recovery (Fig. \ref{RealNMRSpectrum}(d)) of the ground truth (Fig. \ref{RealNMRSpectrum}(a)) than ADM-TR (Fig. \ref{RealNMRSpectrum}(b)) and WCP (Fig. \ref{RealNMRSpectrum}(c)). Moreover, Fig. \ref{RealNMRSpectrum}(d) indicates that HMRTC can achieve high quality of reconstruction even with the sampling rate 10\% and hence allow a significant reduction in measurement time. Thus, the proposed HMRTC algorithm may serve as a versatility method for studying biological or chemical molecules using \emph{N}-D NMR spectroscopy.

%=====================================================
%-------------------- Discussions --------------------
%=====================================================
\section{Discussions}\label{sec:discussion}
In the following, a RLNE smaller than $0.1$ is considered as a low reconstruction error since the corresponding energy loss is less than $1\%$.

% -------------------- 5-D --------------------------
\subsection{Higher-dimensional experiments}\label{sec:high-dimension}
\begin{figure}[htb]
\centering
\includegraphics[width=3.4in]{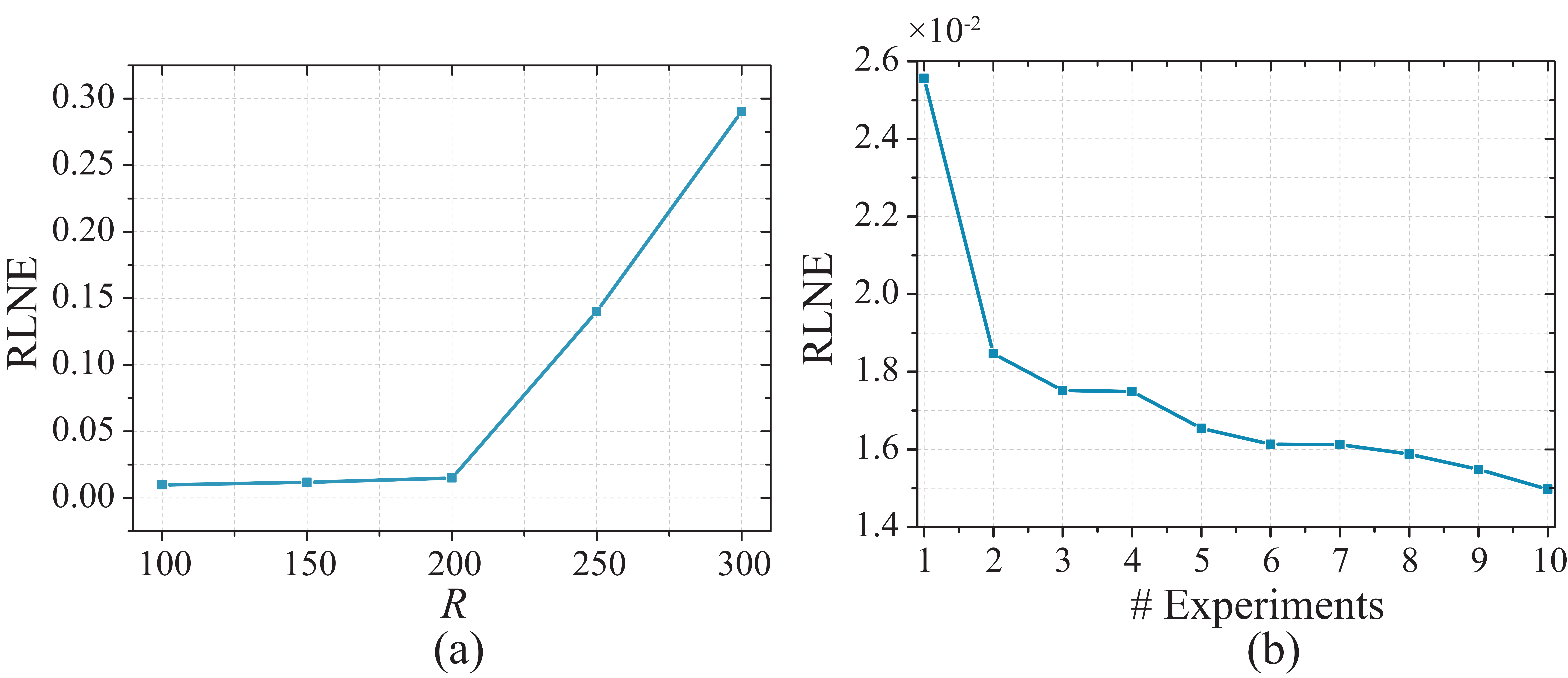}
\caption{(a) The reconstruction error by HMRTC on $5$-D data with the size $30^5$ and sampling rate $SR=0.5$. (b) The effect of different initializations on the $3$-D data in Fig. \ref{RecSpec_MissingSlicesTensor} with sampling ratio  $SR=0.3$. We sort the RLNE for better visualization though these RLNEs are randomly distributed in experiments.}
\label{Discussion_High_dimension and initial value}
\end{figure}

A $5$-D experiment is conducted to explore the capability of HMRTC in higher-dimensional exponential signal recovery. Fig. \ref{Discussion_High_dimension and initial value}(a) indicates that HMRTC achieves a low reconstruction error although $R$ is as large as around $6$ times of ambient dimensions. The reconstruction error will increase dramatically when $R$ further increases ($R \ge 250$). Therefore, HMRTC has the potential to reconstruct the signal with a quite larger $R$ than ambient dimensions. However, reconstructing a signal with larger $R$ needs further development.

%-------------------- other two methods ---------------------
\subsection{Comparison with other state-of-the-art methods}
\begin{figure}[htb]
\centering
\includegraphics[width=3.2in]{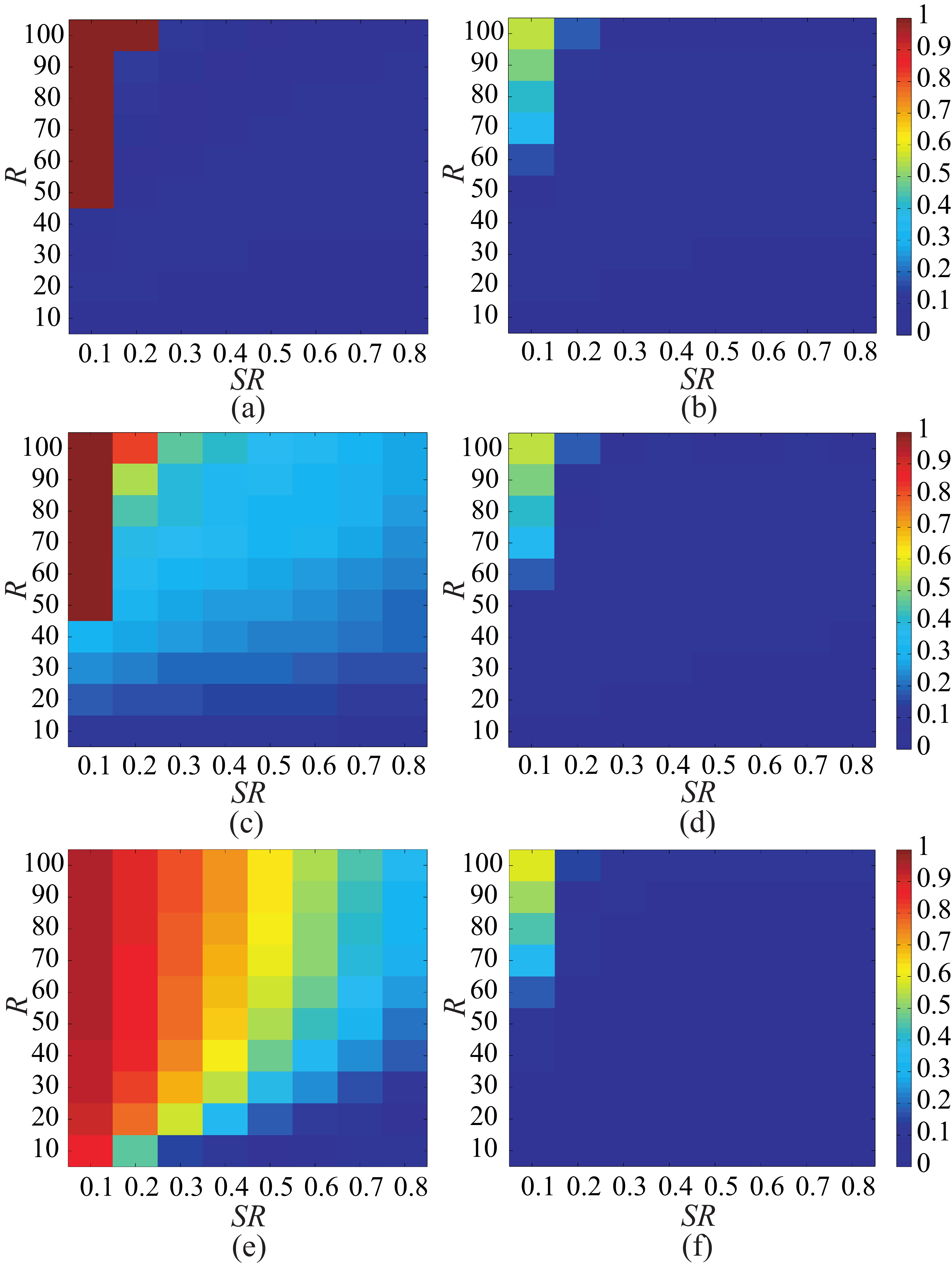}
\caption{Comparison between HMRTC with other state-of-the-art methods. (a) and (b) show average RLNEs by TNN \cite{TNN} and HMRTC for real exponential signal recovery, respectively; (c) and (d) indicate average RLNEs by ADMM-R \cite{ADMM_R} and HMRTC for complex exponential signal recovery, respectively; (e) and (f) show average RLNEs by FaLRTC \cite{sue31} and HMRTC for noiseless complex exponential signal recovery. The experiment settings are based on the consideration that TNN \cite{TNN} cannot support complex number and FaLRTC \cite{sue31} aims to recover noiseless tensors.}
\label{other_methods}
\end{figure}

Another three state-of-the-art methods, ADMM-R \cite{ADMM_R}, TNN \cite{TNN} and FaLRTC \cite {sue31}, are compared with the proposed HMRTC on undamped complex sinusoid recovery. Since tensor nuclear norm is computationally intractable, the approximate solution was proposed recently in \cite{TNN} by providing sub-optimality guarantees. More recently, another computational method to equally achieve tensor nuclear norm was proposed in a preprint paper \cite{ADMM_R}. FaLRTC \cite {sue31} is another typical method in tensor completion which minimizes low $n$-rank.
Fig. \ref{other_methods} shows that HMRTC holds advantage over ADMM-R \cite{ADMM_R}, TNN \cite{TNN} and FaLRTC \cite {sue31} on achieving much lower reconstruction errors. Note that the estimated tensor rank $\hat R$ is set to be exactly ground-truth number of exponentials in ADMM-R \cite{ADMM_R}, which may be unknown in practice.

% --------------------------- Accuracy of recovered factor vectors -------------------------
\subsection{Success rate of factor vector recovery}
The low-CP-rank completion method may lead the factor vector to be consisted of single or multiple exponentials in real applications \cite{MUNIN}. Considering the proposed model also explores the low-CP-rank structure, one may wonder the possibility of the reconstructed factor vectors being a single exponential as it was introduced in \eqref{eq:model}. Here we discuss this possibility based on numerical experiments.

Following \cite{sue51}, the factor vector recovery is declared successful if
\begin{equation*}
{\rm{sim}}\left( r \right) = \frac{{\left| {{\bf{a}}_r^H{{{\bf{\hat a}}}_r}} \right|}}{{\left\| {{{\bf{a}}_r}} \right\|\left\| {{{{\bf{\hat a}}}_r}} \right\|}} \times \frac{{| {{\bf{b}}_r^H{{{\bf{\hat b}}}_r}}|}}{{\left\| {{{\bf{b}}_r}} \right\|\| {{{{\bf{\hat b}}}_r}} \|}} \times \frac{{\left| {{\bf{c}}_r^H{{{\bf{\hat c}}}_r}} \right|}}{{\left\| {{{\bf{c}}_r}} \right\|\left\| {{{{\bf{\hat c}}}_r}} \right\|}} > {0.99^3}
\end{equation*}
for all $r \in \left\{ {1, \ldots ,R} \right\}$,  where ${{\bf{a}}_r}$ is the recovery of the ground truth ${\bf{a}}$. Here the empirical success rate is calculated by averaging over $10$ Monte Carlo trials.

\begin{figure}[htb]
\centering
\includegraphics[width=2in]{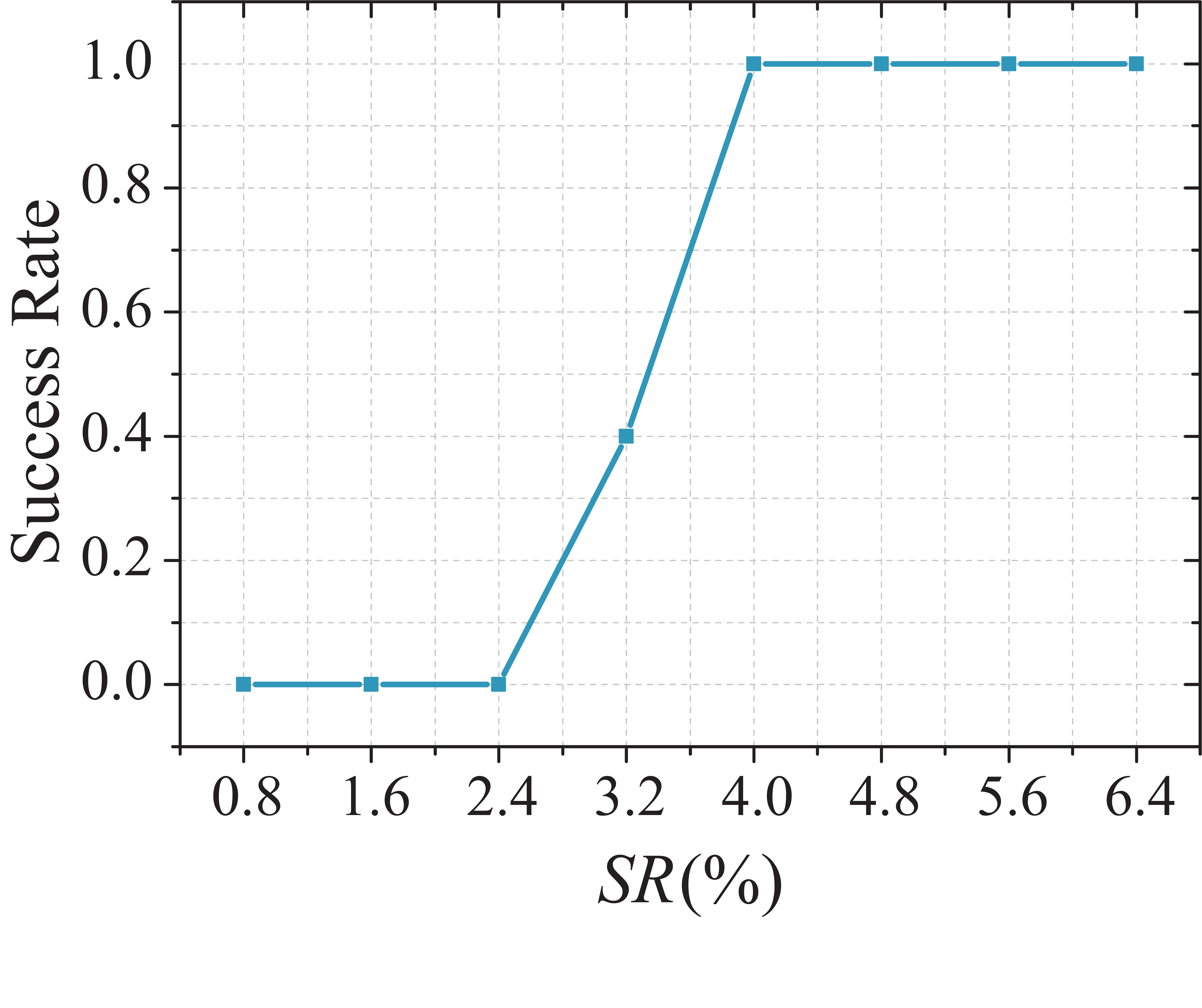}
\caption{The empirical success rate of the factor vector recovery versus sampling ratio. The simulated signals contains 10 exponentials. We set $\hat{R}=10$ and $\rho=1.02$, and measurements are randomly sampled.}
\label{SuccessRate_FactorMatrixRecovery}
\end{figure}

Fig. \ref{SuccessRate_FactorMatrixRecovery} indicates that HMRTC has a high probability to achieve accurate factor vectors, as long as the sampling ratio is sufficiently large, though it is not theoretically guaranteed.

% ------------------- Estimation of Frequency --------------
\subsection{Frequency estimation}
Although the focus of this work lies in reconstructing a full $\textit{N}$-D exponential signal from partial measurements, it is still important to estimate frequency in some applications. Here, we adopt the method in \cite{EstFreq} to estimate the frequency and root mean square error (RMSE) to quantify the accuracy of estimation. The RMSE is defined as
\begin{equation*}
{\textrm{RMSE}} = \sqrt {\frac{1}{R}\sum\limits_{r = 1}^R {{{\left( {{f_r} - {{\hat f}_r}} \right)}^2}} },
\end{equation*}
where ${f_r}$ denotes the estimated frequency of the fully sampled signal, ${\hat f_r}$ the estimated frequency of the reconstructed signal, and $R$ the number of exponentials.
Table \ref{Frequency_Estimation} indicates that the proposed HMRTC obtains the best frequency estimation, comparing with ADM-TR and WCP.

\begin{table*}[tbp]
\centering
\caption{Frequency estimation from different reconstructions.}
\label{Frequency_Estimation}
\begin{tabular}{|c|c|c|c|c|}
\hline
\multirow{2}{*}{\# Peaks} & \multirow{2}{*}{\tabincell{c}{Estimated frequencies \\ from noiseless full data}} & \multicolumn{3}{c|}{Frequency errors ($10^{-5}$)}        \\ \cline{3-5}
                  &                                                                 & ADM-TR               & WCP        & Proposed \\ \hline
1                 & (0.12201, 0.45297, 0.31506)                                     & (Failed,108, 338)    & (9, 1, 5)  & (3,2,2)  \\ \hline
2                 & (0.23503, 0.54308, 0.38204)                                     & (Failed, 2913, 579)  & (1, 5, 20) & (1,1,1)  \\ \hline
3                 & (0.26608, 0.57491, 0.44104)                                     & (Failed, 7207, 3499) & (2, 2, 1)  & (2,0,1)  \\ \hline
4                 & (0.31721, 0.64294, 0.47386)                                     & (Failed, 8213 ,7507) & (6, 14, 2) & (1,0,0)  \\ \hline
5                 & (0.38207, 0.72565, 0.55160)                                     & (252, 1471, 3559)    & (5, 0, 1)  & (2,2,3)  \\ \hline
6                 & (0.41720, 0.74228, 0.58575)                                     & (Failed, 7016, 418)  & (2, 4, 0)  & (2,1,0)  \\ \hline
7                 & (0.44421, 0.81311, 0.62796)                                     & (Failed, 3299, 6558) & (0, 4, 3)  & (1,1,0)  \\ \hline
8                 & (0.56599, 0.84617, 0.69391)                                     & (Failed, 5922, 4492) & (3, 0, 3)  & (0,1,2)  \\ \hline
9                 & (0.62214, 0.90486, 0.73689)                                     & (235, 3533, 6835)    & (6, 1,3)   & (0,0,1)  \\ \hline
10                & (0.69392, 0.94303, 0.81405)                                     & (51, 105, 388)       & (2, 0, 3)  & (1,0,1)  \\ \hline
\end{tabular}
\begin{tablenotes}
  \footnotesize
  \item Note: The signal reconstruction error, RLNE, of ADM-TR, WCP and the proposed method are 0.9321, 0.0373 and 0.0105, respectively. The $\textit{Failed}$ means that too many pseudo peaks are presented thus the frequency cannot be estimated. The noisy $3$-D signal with known frequencies in Fig. \ref{RecSpec_MissingSlicesTensor} is used for simulation. The number of peaks $R$ is 10 and the sampling rate is 6\%.
\end{tablenotes}
\end{table*}

% ---------------------- Discussion on Parameter Setting ----------------
\subsection{Regularization parameter}\label{sec:parameter}
\begin{figure}[htb]
\centering
\includegraphics[width=3.4in]{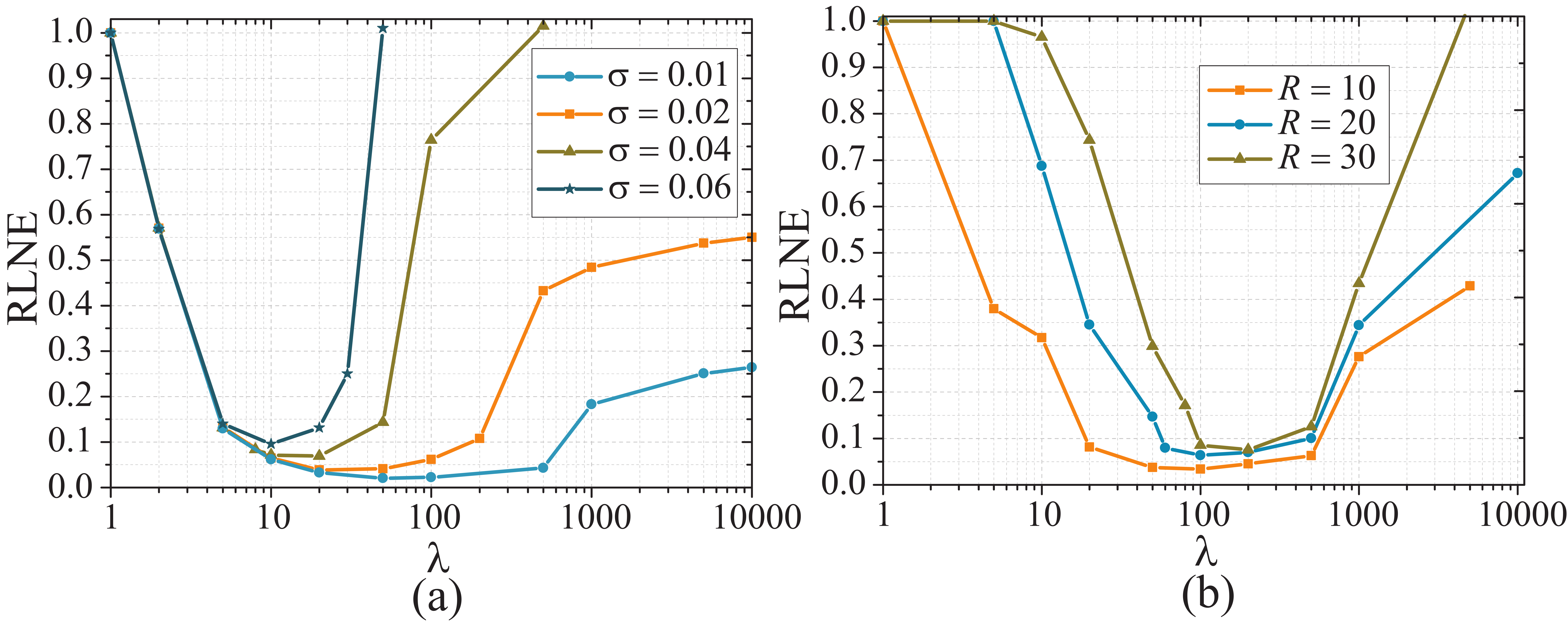}
\caption{ Reconstruction errors versus regularization parameter $\lambda$. (a) denotes RLNE via $\lambda$ under different noise level $\sigma$, (b) denotes RLNE via $\lambda$ under different $R$. Note: The source $3$-D data in (a) is Fig. \ref{RecSpec_MissingSlicesTensor} and in (b) is generated with noise level 0.01. The sampling ratio in (a) and (b) is $0.3$. }
\label{Lambda_Sigma_R}
\end{figure}

The optimal $\lambda $ in HMRTC, producing the lowest reconstruction error, generally decreases as the noise level increases. For example, as shown in Fig. \ref{Lambda_Sigma_R}(a), the optimal $\lambda $ is $50$, $20$ and $10$ when the noise level is $0.01$, $0.02$ and $0.04$, respectively. This trend means that a smaller $\lambda $ should be set for a higher noise level.

\begin{figure}[htb]
\centering
\includegraphics[width=3.1in]{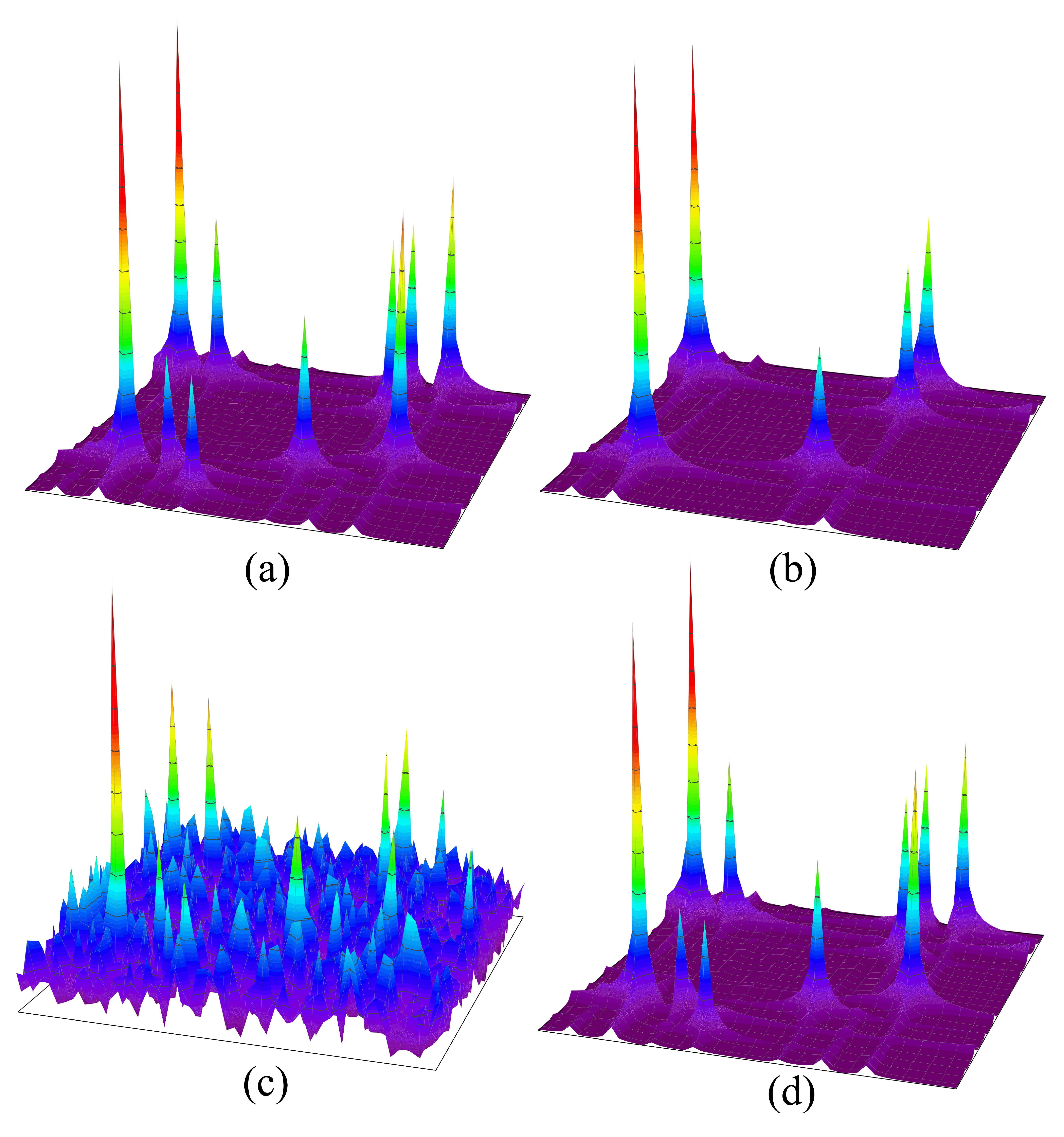}
\caption{A $2$-D plane of the reconstructed $3$-D spectrum with different regularization parameters $\lambda $. (a) is the noiseless ground-truth spectrum, (b)-(d) are reconstructed spectra with $\lambda  = 2$, $1000$ and $50$, respectively. Note: The $3$-D data in Fig. \ref{RecSpec_MissingSlicesTensor} is used for simulation. The sampling ratio $SR=0.3$, the number of exponentials $R=10$ and the noise level $\sigma  = 0.02$.}
\label{Sigma_002_RecSpectrum}
\end{figure}

Fig. \ref{Lambda_Sigma_R}(a) indicates that there exists an available range ($10 \le \lambda  \le {10^2}$) for $\lambda $ that leads to low reconstruction errors (${\rm{RLNE}} \le 0.1$), while a smaller or larger $\lambda$ will bring high reconstruction errors (${\rm{RLNE}} > 0.1$). It is primarily because that a smaller $\lambda $ results in missing some spectral peaks (Fig. \ref{Sigma_002_RecSpectrum}(b)) while a larger $\lambda $ introduces more noise into the spectrum (Fig. \ref{Sigma_002_RecSpectrum}(c)). In addition, this available range turns narrowed as the noise level increases (Fig. \ref{Lambda_Sigma_R}(a)). Fig. \ref{Lambda_Sigma_R}(b) further indicates that, the available range of $\lambda$ turns narrowed if the number of exponentials increases, implying that it becomes more difficult to find an optimal $\lambda$.

% ---------------------- Initialization -------------------
\subsection{Effect of local minima}
Due to the CP decomposition, HMRTC is non-convex and thus local minima may be encountered. Fig. \ref{Discussion_High_dimension and initial value}(b) presents that, under different initializations, all reconstruction errors RLNEs are very small (in the order of ${10^{ - 2}}$) though each reconstruction error varies. We further quantitatively compare the frequency estimation errors of reconstructions with the maximum and minimum RLNE in Fig. \ref{Discussion_High_dimension and initial value}(b). Table \ref{EstFreq_Init} shows that each frequency estimation error is in the order of ${10^{ - 5}}$, implying that the initializations has no significant impact on the reconstruction.

%\begin{figure}[htb]
%\centering
%\includegraphics[width=2.6in]{AlgriInit_RLNE.png}
%\caption{The RLNE of reconstructions with different initializations. Note: The 3D data in Fig. \ref{RecSpec_MissingSlicesTensor} that contains 10 peaks is used for simulation and the sampling rate is 0.3. We sort the RLNE for better visualization though these RLNE are randomly distributed in experiments.}
%\label{AlgriInit_RLNE}
%\end{figure}

\begin{table}[tbp]
\centering
\caption{Comparisons on frequency estimation error of reconstructions using different initializations.}
\label{EstFreq_Init}
\begin{tabular}{|c|p{3.5cm}<{\centering}|c|c|}
\hline
\multirow{2}{*}{\# Peaks} & \multirow{2}{*}{\centering \tabincell{c}{Frequencies estimation \\from noiseless full data}} & \multicolumn{2}{c|}{\tabincell{c}{Frequency estimation \\ errors ($10^{-5}$)}} \\ \cline{3-4}
                  &                                                                  & \tabincell{c}{Maximum \\ RLNE}        & \tabincell{c}{Minimum \\ RLNE}       \\ \hline
1                 & (0.69999, 0.29999, 0.30005)                                      & (1, 0, 2)                  & (1, 13, 2)                 \\ \hline
2                 & (0.19994, 0.40000, 0.70007)                                      & (2, 2, 2)                  & (4, 1, 0)                  \\ \hline
3                 & (0.29997, 0.7999, 0.59999)                                       & (2, 7, 2)                  & (1, 9, 3)                  \\ \hline
4                 & (0.80004, 0.80007, 0.79944)                                      & (1, 2, 1)                  & (1, 3, 1)                  \\ \hline
5                 & (0.09992, 0.90001, 0.20001)                                      & (1, 3, 0)                  & (1, 4, 1)                  \\ \hline
6                 & (0.20178, 0.69992, 0.09993)                                      & (3, 2, 2)                  & (11, 2, 3)                 \\ \hline
7                 & (0.19992, 0.30194, 0.69963)                                      & (0, 4, 1)                  & (1, 26, 3)                 \\ \hline
8                 & (0.30004, 0.70162, 0.59979)                                      & (0, 0, 0)                  & (1, 2, 1)                  \\ \hline
9                 & (0.89893, 0.20247, 0.29986)                                      & (7, 0, 1)                  & (6, 0,0)                   \\ \hline
10                & (0.29743, 0.10002, 0.39997)                                      & (4, 5, 0)                  & (4, 4, 0)                  \\ \hline
\end{tabular}
\begin{tablenotes}
  \footnotesize
  \item Note: The frequency estimation are conducted on reconstructions with the maximum and minimum RLNE in Fig. \ref{Discussion_High_dimension and initial value}(b).
\end{tablenotes}
\end{table}

% ===============================================================
% ----------------- Conclusion and future work ------------------
% ===============================================================
\section{Conclusions and future work}\label{sec:conclusion}
A Hankel matrix nuclear norm regularized CANDECOMP/PARAFAC(CP)-based tensor completion method is proposed to recover missing data of high-dimensional exponential signals. The full signal is reconstructed by simultaneously exploring the low-CP-rank structure of the signal and the exponential structure of the associated factor vectors. Numerical experiments on simulated undamped and damped complex sinusoids, as well as the real NMR protein spectroscopy data, demonstrate that the proposed method can successfully recover full signals from very limited samples and is robust to the estimated tensor rank. The proposed method is particularly useful for fast sampling of high-dimensional spectroscopy in chemistry, biology and imaging sciences, especially those applications where the number of spectrum peaks is hard to estimate.

The proposed algorithm may face challenge in computation when the number of dimensions and ambient dimensions are large and thus developing a more efficient numerical algorithm to accommodate larger datasets will be considered in the further work. For example, it is meaningful to consider more computationally appealing characterization of the nuclear norm \cite{MP_2005, LT_2005, TSP_2015} to accelerate the algorithm. In addition, it is worthy of deriving the recovery conditions of this method, including the possible frequency separation conditions.
% ===============================================================
% ----------------------- Acknowledgments -----------------------
% ===============================================================
\section{Acknowledgments}\label{sec:Acknowledgments}
The authors would like to thank Silvia Gandy, Gongguo Tang, Ji Liu, Xinhua Zhang and Andreas Jakobsson for sharing codes for comparisons and Weiyu Xu for helpful discussions. The authors also appreciate reviewers and editors for their constructive comments.

% ===============================================================
% -------------------------- appendix ---------------------------
% ===============================================================
\begin{appendix}

% ------------------ Derivation of closed form solution of U^n --------------------
\subsection{Derivation of closed form solution of $\bf{U}^{(n)}$}
1) Operator Definition

Before deriving the closed-form of ${\bf{U}}^{\left( n \right)}$, we define two linear operators and their adjoint operators.
We rewrite the definition of Hankel operator ${{\cal R}}$ as follows
\begin{equation}
\begin{aligned}
& {{\cal R}}: {\bf{x}} \in {\mathbb{C}^{{s_1} + {s_2} - 1}} \mapsto  {{\cal R}}{\bf{x}} \in {\mathbb{C}^{{s_1} \times {s_2}}}, {[{{\cal R}}{\bf{x}}]_{i,j}} = {x_{i + j}}, \\
& \forall i \in \{ 0, \ldots ,{s_1}{\rm{ - }}1\}, j \in \{ 0, \ldots ,{s_2} - 1\},
\end{aligned}
\end{equation}

Meanwhile, the adjoint ${{{\cal R}}^*}$ of ${{\cal R}}$ is given by
\begin{equation}
\begin{aligned}
& {{{\cal R}}^*}: {\kern 1pt} {\kern 1pt} {\bf{X}} \in {\mathbb{C}^{{s_1} \times {s_2}}} \mapsto {{{\cal R}}^*}{\bf{X}} \in {\mathbb{C}^{{s_1} + {s_2} - 1}}, \\
& {[{{{\cal R}}^*}{\bf{X}}]_k} = \sum\nolimits_{i + j = k} {{x_{i,j}}}, \\
& \forall i \in \{ 0, \ldots ,{s_1}{\rm{ - }}1\}, j \in \{ 0, \ldots ,{s_2}{\rm{ - }}1\},
\end{aligned}
\end{equation}
where ${[  \cdot  ]_k}$ and ${[  \cdot  ]_{i,j}}$ denote the \textit{k}-th entry of a vector and the $(i, j)$-element of a matrix, respectively.

We denote ${{\cal D}} = {{{\cal R}}^*}{{\cal R}}$ and it is a diagonal operator from a vector to a vector of the form. In other words, for any ${\bf{x}} \in {\mathbb{C}^{{s_1} + {s_2} - 1}}$, we have
\begin{equation}\label{eq:OP_1}
{{\cal D}}{\bf{x}} = {\bf{w}} \circledast {\bf{x}},
\end{equation}
where $\bf{w}$ is a vector and its \textit{k}-th element ${w_k}$ is the number of elements in \textit{k}-th anti-diagonal of a matrix with size ${s_1} \times {s_2}$; $\circledast$ denotes the Hadamard product.

We also define another linear operator ${{{\cal Q}}_r}$ which aims to extract the $r^{\textrm{th}}$ column from $\mathbf{X}$. For a matrix ${\bf{X}} \in {\mathbb{C}^{{n_1} \times {n_2}}}$, specifically, we define ${{{\cal Q}}_r}$ by
\begin{equation}
\begin{aligned}
& {{{\cal Q}}_r}: {\kern 1pt} {\bf{X}} \in {\mathbb{C}^{{n_1} \times {n_2}}} \mapsto {{{\cal Q}}_r}{\bf{X}} = {{\bf{X}}_{(:,r)}} \in {\mathbb{C}^{{n_1} \times 1}}, \\
& \forall r \in \{ 0, \ldots ,{n_2}{\rm{ - }}1\}.
\end{aligned}
\end{equation}

Then the adjoint ${{\cal Q}}_r^*$ of ${{{\cal Q}}_r}$ is given by
\begin{equation}
\begin{aligned}
& {{\cal Q}}_r^*: {\bf{x}} \in {\mathbb{C}^{{n_1} \times 1}} \mapsto {{\cal Q}}_r^*{\bf{x}} \in {\mathbb{C}^{{n_1} \times {n_2}}}, \\
& {\left[ {{{\cal Q}}_r^*{\bf{x}}} \right]_{(:,k)}} = \left\{ {
\begin{array}{*{20}{c}}
{\bf{x}}\\
0
\end{array}
\begin{array}{*{20}{c}}
{}\\
{}
\end{array}
\begin{array}{*{20}{c}}
{k = r,}\\
{k \ne r.}
\end{array}} \right. \forall r \in \{ 0, \ldots ,{n_2}{\rm{ - }}1\}.
\end{aligned}
\end{equation}
where ${\left[  \cdot  \right]_{(:,k)}}$ denotes the \textit{k}-th column of a matrix, $k = 0, \ldots ,{s_2}{\rm{ - }}1$.Thus we have
\begin{equation}\label{eq:OP_2}
{\left[ {{{\cal Q}}_r^*{{\cal Q}}_r^{}{\bf{X}}} \right]_{(:,k)}} = \left\{ {
\begin{array}{*{20}{c}}
{{{\bf{X}}_{(:,r)}}}\\
0
\end{array}
\begin{array}{*{20}{c}}
{k = r,}\\
{k \ne r.}
\end{array}} \right.
\end{equation}

According to \eqref{eq:OP_1} and \eqref{eq:OP_2}, for any matrix ${\bf{X}} \in {\mathbb{C}^{{n_1} \times {n_2}}}$, we have
\begin{equation}\label{eq:OP_3}
\sum\limits_{r = 1}^{{n_2}} {{{{\cal Q}}_r}{{\cal R}{\cal R}}{{{\cal Q}}_r}{\bf{X}}}  = {\bf{C}} \circledast {\bf{X}},
\end{equation}
where ${\bf{C}} \in {\mathbb{C}^{{n_1} \times {n_2}}}$ and each column equals to $\mathbf{w}$ defined in \eqref{eq:OP_1}. Note that here ${n_1} = {s_1} + {s_2} - 1$, i.e., Hankel operator ${{\cal R}}$ maps a vector with size ${n_1}$ to a Hankel matrix with size ${s_1} \times {s_2}$.
%\begin{equation}\label{eq:OP_4}
%\begin{aligned}
%& \mathop {{\rm{min}}}\limits_{\substack{{{\bf{U}}^{( n )}}\\ n=1,\ldots,N} }
%\sum\limits_{r = 1}^{\hat R} {\sum\limits_{n = 1}^N {{{\big\| {{{\cal R}}{Q_r}{{\bf{U}}^{(n)}}} \big\|}_*}}} + \\
%& {\frac{\lambda }{2}\big\| {{{{\cal P}}_\Omega }({{\cal Y}}) - {{{\cal P}}_\Omega }(\big[\kern-0.25em\big[ {{{\bf{U}}^{(1)}},{{\bf{U}}^{(2)}}, \ldots ,{{\bf{U}}^{(N)}}} \big]\kern-0.25em\big])} \big\|_F^2},
%\end{aligned}
%\end{equation}

2) Derivation of closed form solution of $\bf{U}^{(n)}$

For simplicity, we replace the terms in the right of \eqref{eq:ADMM_5} with a constant matrix ${\bf{E}}{\rm{ = }}\lambda {\cal A}_k^{\left( n \right)*}\left( {{{\cal P}_{{\Omega ^{\left( n \right)}}}}\left( {{{\bf{Y}}_{\left( n \right)}}} \right)} \right) + {\beta _k}\sum\limits_{r = 1}^{\hat R} {{\cal Q}_r^*{{\cal R}^*}\left( {{\bf{Z}}_{r;k}^{\left( n \right)} - {{\left( {{\beta _k}} \right)}^{ - 1}}{\bf{D}}_{r;k}^{\left( n \right)}} \right)} $, and by \eqref{eq:OP_3} one has
\begin{equation}\label{closed_E}
\lambda {{\cal A}}_k^{\left( n \right)*}{{\cal A}}_k^{\left( n \right)}{{\bf{U}}^{\left( n \right)}} + \beta_{k} {\bf{C}} \circledast {{\bf{U}}^{\left( n \right)}} = {\bf{E}}.
\end{equation}
Since ${{\cal A}}_k^{\left( n \right)}$ can be rewritten as
\begin{equation}
\begin{aligned}
& {{\cal A}}_k^{\left( n \right)}\left( {{{\bf{U}}^{\left( n \right)}}} \right) = {{\cal P}}_{{\Omega ^{\left( n \right)}}}^{}\left( {{\kern 1pt} {{\bf{U}}^{\left( n \right)}}{\bf{G}}_k^{\left( n \right)}} \right) \\
& = \left[ {
\begin{array}{*{20}{c}}
{{{\cal P}}_{\Omega _1^{\left( n \right)}}^{}\left( {{\bf{v}}_1^{\left( n \right)}{\bf{G}}_k^{\left( n \right)}} \right)}\\
 \vdots \\
{{{\cal P}}_{\Omega _{{I_n}}^{\left( n \right)}}^{}\left( {{\bf{v}}_{{I_n}}^{\left( n \right)}{\bf{G}}_k^{\left( n \right)}} \right)}
\end{array}} \right] = \left[ {
\begin{array}{*{20}{c}}
{{{\cal A}}_{1{\rm{;}}k}^{\left( n \right)}\left( {{\bf{v}}_1^{\left( n \right)}} \right)}\\
 \vdots \\
{{{\cal A}}_{{I_n}{\rm{;}}k}^{\left( n \right)}\left( {{\bf{v}}_{{I_n}}^{\left( n \right)}} \right)}
\end{array}} \right],
\end{aligned}
\end{equation}
where ${\bf{v}}_i^{\left( n \right)} \in {\mathbb{C}^{1 \times \hat R}}$ denotes the $i$-th row of ${{\bf{U}}^{\left( n \right)}}$ and $\Omega _i^{\left( n \right)}$ is the subset of $\Omega ^{\left( n \right)}$ corresponding the $i$-th row. Here ${{\cal A}}_{i{\rm{;}}k}^{\left( n \right)}\left( {\bf{x}} \right) \buildrel \Delta \over = {{\cal P}}_{\Omega _i^{\left( n \right)}}\left( {{\bf{xG}}_k^{\left( n \right)}} \right)$. Therefore, to obtain the closed-form of ${{\bf{U}}^{\left( n \right)}}$, we can first derive the closed-form of each row of ${{\bf{U}}^{\left( n \right)}}$, i.e., ${\bf{v}}_i^{\left( n \right)}$. Then \eqref{closed_E} can be written as
\begin{equation}\label{eq_40}
\begin{aligned}
&\lambda {{\cal A}}_{i{\rm{;}}k}^{\left( n \right)*}{{\cal A}}_{i{\rm{;}}k}^{\left( n \right)}{\rm{(}}{\bf{v}}_i^{\left( n \right)}{\rm{)}} + \beta {{\bf{c}}_i} \circledast {\bf{v}}_i^{\left( n \right)} = {{\bf{e}}_i} \\
& \qquad \qquad {\rm{for}} \: i \in \{ 1, \cdots ,{I_n}\},
\end{aligned}
\end{equation}
where ${{\bf{e}}_i}$ and ${{\bf{c}}_i}$ denote the $i$-th row of $\mathbf{E}$ and $\mathbf{C}$, respectively.

To get the closed form of ${\bf{v}}_i^{\left( n \right)}$, we simplify the operator ${{\cal A}}_{i{\rm{;}}k}^{\left( n \right)}$. According to the definition ${{\cal A}}_{i{\rm{;}}k}^{\left( n \right)}\left( {\bf{x}} \right) = {{\cal P}}_{\Omega _i^{\left( n \right)}}^{}\big( {{\bf{xG}}_k^{\left( n \right)}} \big)$ and ${\bf{xG}}_k^{\left( n \right)} \in {\mathbb{C}^{1 \times {K_n}}}$ with ${K_n} = \prod\nolimits_{k = 1,k \ne n}^N {{I_k}} $, ${{\cal A}}_{i{\rm{;}}k}^{\left( n \right)}$ can be rewritten as ${{\cal A}}_{i{\rm{;}}k}^{\left( n \right)}\left( {\bf{x}} \right) = {\bf{xG}}_k^{\left( n \right)}{{\bf{P}}_i}$, where ${{\bf{P}}_i} \in {\mathbb{C}^{{K_n} \times {K_n}}}$is a diagonal matrix and the $\left( {k,k} \right)$-th entry in the main diagonal is
\begin{equation}
{[{{\bf{P}}_i}]_{k,k}} = \left\{ {
\begin{array}{*{20}{c}}
\begin{aligned}
&{ 1 \quad {\rm{if}}{\kern 1pt}(k) \in \Omega _i^{\left( n \right)},}\\
&{ 0 \quad {\rm{otherwise}}.}
\end{aligned}
\end{array}} \right.
\end{equation}
for all $k \in \{ 1, \ldots ,{K_n}\} $, where ${[ \cdot ]_{k,k}}$ denotes the $\left( {k,k} \right)$-th entry of a matrix. Hence \eqref{eq_40} can be written as
\begin{equation}
\lambda {\bf{v}}_i^{\left( n \right)}{\bf{G}}_k^{\left( n \right)}{{\bf{P}}_i}{\bf{P}}_i^*{\bf{G}}_k^{\left( n \right)*} + \beta_{k} {\bf{v}}_i^{\left( n \right)}D\left( {{{\bf{c}}_i}} \right) = {{\bf{e}}_i},
\end{equation}
where $D\left( {{{\bf{c}}_i}} \right)$ denotes a diagonal matrix and the $\left( {k,k} \right)$-th entry in $D\left( {{{\bf{c}}_i}} \right)$ equals to the $\textit{k}$-th entry in ${{\bf{c}}_i}$.

Therefore the closed form of ${\bf{v}}_i^{\left( n \right)}$ is
\begin{equation}
{\bf{v}}_i^{\left( n \right)} = {{\bf{e}}_i}{\left( {\lambda {\bf{G}}_k^{\left( n \right)}{{\bf{P}}_i}{\bf{P}}_i^*{\bf{G}}_k^{\left( n \right)*} + \beta_{k} D\left( {{{\bf{c}}_i}} \right)} \right)^{ - 1}}.
\end{equation}

Then the closed form solution of $\bf{U}^{(n)}$ can be obtained by ${{\bf{U}}^{\left( n \right)}} = {[{\bf{v}}_1^{\left( n \right)T} \ldots {\bf{v}}_{{I_n}}^{\left( n \right)T}]^T}$
% ----------------------- Proof of lemmas -----------------------------
\newtheorem*{proof}{Proof}
\newtheorem{lemma}{Lemma}

\subsection{Lemmas}
% lemma 1
To prove \textbf{Theorem 1} and $\mathbf{2}$, we first prove the boundedness of multipliers and some variables generated by Algorithm \ref{alg:1}, and then we analyze the convergence of the algorithm.
\begin{lemma}
\cite{ma} Let ${{\cal T}}$ be Hilbert space endowed with an inner product $\left\langle { \cdot , \cdot } \right\rangle $ and a corresponding norm $\left\|  \cdot  \right\|$, and ${\bf{y}} \in \partial \left\| {\bf{x}} \right\|$, where $\partial f({\bf{x}})$ is the subgradient of a convex $f({\bf{x}})$. Then ${\left\| {\bf{y}} \right\|^*} = 1$ if ${\bf{x}} \ne {\bf{0}}$, and ${\left\| {\bf{y}} \right\|^*} \le 1$ if ${\bf{x}} = {\bf{0}}$, where ${\left\|  \cdot  \right\|^*}$ is the dual norm of $\left\|  \cdot  \right\|$.
\end{lemma}
% lemma 2
\begin{lemma}
The sequences $\{ {\bf{D}}_{r;k}^{\left( n \right)}\} $, $r = 1, \ldots ,\hat R$ and $n = 1, \ldots ,N$, are bounded.
\end{lemma}

% proof of lemma 2
\noindent \textit{Proof:}
By the optimality of ${\bf{Z}}_{r;k + 1}^{\left( n \right)}$, we have that:
$$
{\bf{0}} \in \partial {\big\| {{\bf{Z}}_{r;k + 1}^{\left( n \right)}} \big\|_*} - {\bf{D}}_{r;k}^{\left( n \right)} - \beta_k\left({{\cal R}}{{{\cal Q}}_r}{\bf{U}}_k^{\left( n \right)} - {\bf{Z}}_{r;k + 1}^{\left( n \right)}\right).
$$
This together with
$$
{\bf{D}}_{r;k + 1}^{\left( n \right)}={\bf{D}}_{r;k}^{\left( n \right)} + \beta_k\left({{\cal R}}{{{\cal Q}}_r}{\bf{U}}_k^{\left( n \right)} - {\bf{Z}}_{r;k + 1}^{\left( n \right)}\right)
$$
implies
$$
{\bf{D}}_{r;k + 1}^{\left( n \right)} \in \partial {\big\| {{\bf{Z}}_{r;k + 1}^{\left( n \right)}} \big\|_*}.
$$

From \textbf{lemma 1} and \cite{sue42}, every element of the subgradient of the nuclear norm is bounded by 1 in spectral norm. Therefore, ${\big\| {{\bf{D}}_{r;k}^{\left( n \right)}} \big\|_2} \le 1$ and hence the sequence $\{ {\bf{D}}_{r;k}^{\left( n \right)}\} $ is bounded for all $r$ and $n$.

% lemma 3
\begin{lemma}
The sequences $\{ {\bf{U}}_k^{\left( n \right)}\} $ and $\{ {\bf{Z}}_{r;k}^{\left( n \right)}\} $, $r = 1, \ldots ,\hat R$, $n = 1, \ldots ,N$, produced by algorithm 1 are bounded.
\end{lemma}
% proof of lemma 3
\noindent \textit{Proof:}
From the iteration procedure, we have that
\begin{equation*}
\begin{aligned}
& {{{\cal L}}_{{\beta _k}}}({{{\cal U}}_{k + 1}},{{{\cal Z}}_{k + 1}}, {{{\cal D}}_k}) \le {{{\cal L}}_{{\beta _k}}}({{{\cal U}}_k}, {{{\cal Z}}_k},{{{\cal D}}_k})\\
& = \sum\limits_{n = 1}^N {\sum\limits_{r = 1}^{\hat R} {\big( {\big\langle {{\bf{D}}_{r;k}^{\left( n \right)} - {\bf{D}}_{r;k - 1}^{\left( n \right)},{{\cal R}}{{{\cal Q}}_r}{\bf{U}}_k^{\left( n \right)} - {\bf{Z}}_{r;k}^{\left( n \right)}} \big\rangle }}} +\\
&{{{ \frac{{{\beta _k} - {\beta _{k - 1}}}}{2}\big\| {{{\cal R}}{{{\cal Q}}_r}{\bf{U}}_k^{\left( n \right)} - {\bf{Z}}_{r;k}^{\left( n \right)}} \big\|_F^2} \big)} } + {{{\cal L}}_{{\beta _{k - 1}}}}({{{\cal U}}_k},{{{\cal Z}}_k}, {{{\cal D}}_{k - 1}})\\
& = \sum\limits_{n = 1}^N {\sum\limits_{r = 1}^{\hat R} {\big( { \frac{1}{\beta _{k - 1}} {\big\| {{\bf{D}}_{r;k}^{\left( n \right)} - {\bf{D}}_{r;k - 1}^{\left( n \right)}} \big\|_F^2 } }}} \\
& + {{{ \frac{{{\beta _k} - {\beta _{k - 1}}}}{2\beta _{k - 1}^2}\big\| {{\bf{D}}_{r;k}^{\left( n \right)} - {\bf{D}}_{r;k - 1}^{\left( n \right)}} \big\|_F^2} \big)} } + {{{\cal L}}_{{\beta _{k - 1}}}}({{{\cal U}}_k},{{{\cal Z}}_k}, {{{\cal D}}_{k - 1}})\\
& = \frac{{{\beta _k} + {\beta _{k - 1}}}}{{2{ \beta _{k - 1}^2} }}\sum\limits_{n = 1}^N {\sum\limits_{r = 1}^{\hat R} {\big\| {{\bf{D}}_{r;k}^{\left( n \right)} - {\bf{D}}_{r;k - 1}^{\left( n \right)}} \big\|_F^2 } }  \\
& +{{{\cal L}}_{{\beta _{k - 1}}}}({{{\cal U}}_k}, {{{\cal Z}}_k},{{{\cal D}}_{k - 1}}).
\end{aligned}
\end{equation*}

Therefore ${{{\cal L}}_{{\beta _{k - 1}}}}({{{\cal U}}_k},{{{\cal Z}}_k},{{{\cal D}}_{k - 1}})$ is upper bounded thanks to the boundedness of $\{ {\bf{D}}_{r;k}^{\left( n \right)}\} $ and
\begin{equation*}
\sum\limits_{k = 1}^\infty  {\frac{{\beta _k + \beta _{k - 1}}}{{2{{(\beta _{k - 1})}^2}}} = \frac{{\rho (\rho  + 1)}}{{2\beta _0(\rho - 1)}}}  < \infty,
\end{equation*}
where $\rho  \in (1.0,\, 1.1]$.
Then we have
\begin{equation*}
\begin{aligned}
& \scalebox{0.95}{$\sum\limits_{r = 1}^{\hat R} {\sum\limits_{n = 1}^N {{{\big\| {{\bf{Z}}_{r;k}^{(n)}} \big\|}_*}} } $}\\
& \scalebox{0.95}{${\kern 10pt}+\frac{\lambda }{2}\big\| {{{{\cal P}}_\Omega }({{\cal Y}}) - {{{\cal P}}_\Omega }\big( {\big[\kern-0.25em\big[ {{{\bf{U}}_k^{(1)}},{{\bf{U}}_k^{(2)}}, \ldots ,{{\bf{U}}_k^{(N)}}} \big]\kern-0.25em\big]} \big)} \big\|_F^2 $}\\
=& \scalebox{0.95}{${{{\cal L}}_{{\beta _{k - 1}}}}({{{\cal U}}_k},{{{\cal Z}}_k},{{{\cal D}}_{k - 1}})-\frac{1}{{{2\beta _{k - 1}}}}\sum\limits_{n = 1}^N \sum\limits_{r = 1}^{\hat R} \big({\| {{\bf{D}}_{r;k}^{\left( n \right)} - {\bf{D}}_{r;k - 1}^{\left( n \right)}} \|_F^2} $}\\
& \scalebox{0.95}{$ + 2{\big\langle {{\bf{D}}_{r;k - 1}^{\left( n \right)},{\bf{D}}_{r;k}^{\left( n \right)} - {\bf{D}}_{r;k - 1}^{\left( n \right)}} \big\rangle } \big)$} ,
\end{aligned}
\end{equation*}
is upper bounded, which implies $\{{\bf{Z}}_{r;k}^{(n)}\}$ is bounded for all $r \in \{ 1, \ldots ,\hat R\} $ and $n \in \{ 1, \ldots ,N\} $.
Furthermore, $\{ {\bf{U}}_k^{\left( n \right)}\}$ is bounded for $n = 1, \ldots ,N$, due to the equation ${{\cal R}}{{{\cal Q}}_r}{\bf{U}}_{k + 1}^{\left( n \right)} = {\bf{Z}}_{r;k + 1}^{\left( n \right)} + \big( {{\bf{D}}_{r;k + 1}^{\left( n \right)} - {\bf{D}}_{r;k}^{\left( n \right)}} \big)/\beta _k$ for $r = 1, \ldots ,\hat R$.

% ---------------------------- Proof of convergence -------------------
\subsection{Proof of convergence}
%\newtheorem{theorem_appendix}{Theorem}
%\begin{theorem_appendix}
%The sequences $\big \{ {\bf{U}}_k^{\left( n \right)} \big\} $ for all $n = 1, \ldots ,N$ generated by the ADMM algorithm are Cauchy sequences, where ${\bf{U}}_k^{\left( n \right)}$ is the k-th update of the variable ${\bf{U}}_{}^{\left( n \right)}$.
%\end{theorem_appendix}
%
%\begin{theorem_appendix}
%The accumulation point of the sequence $\big( {\big\{ {{\bf{U}}_k^{\left( 1 \right)}} \big\}, \ldots ,\big\{ {{\bf{U}}_k^{\left( N \right)}} \big\}} \big)$ satisfies the KKT conditions for \eqref{eq:OP_4} when $\mathop {\lim }\limits_{k \to \infty } {\kern 1pt} {\kern 1pt} \big\| {{\bf{D}}_{r;k + 1}^{\left( n \right)} - {\bf{D}}_{r;k}^{\left( n \right)}} \big\|_F^{} = 0$, where ${\bf{D}}_{r;k}^{\left( n \right)}$ is the \textit{k}-th update of the variable ${\bf{D}}_r^{\left( n \right)}$, for $r = 1, \ldots ,\hat R$, $n = 1, \ldots ,N$.
%\end{theorem_appendix}

% proof of theorem 1
\noindent \textit{Proof of Theorem 1:}
By ${\bf{D}}_{r;k + 1}^{\left( n \right)} = {\bf{D}}_{r;k}^{\left( n \right)} + \beta _k({{\cal R}}{{{\cal Q}}_r}{\bf{U}}_{k + 1}^{\left( n \right)} - {\bf{Z}}_{r;k + 1}^{\left( n \right)})$ , the boundedness of $\{ {\bf{D}}_{r;k}^{\left( n \right)}\} $ and ${\lim _{k \to \infty }}\beta _k = \infty$, we have
\begin{equation}\label{theorem_1_1}
\lim \limits_{k \to \infty } {\left\| {{{\cal R}}{{{\cal Q}}_r}{\bf{U}}_{k + 1}^{\left( n \right)} - {\bf{Z}}_{r;k + 1}^{\left( n \right)}} \right\|_F} = 0,
\end{equation}
for $r = 1, \ldots ,\hat R$ and $n = 1, \ldots ,N$. Therefore, $\{ {\cal {Z}}_k,{\cal {U}}_k\}$ approaches to a feasible solution.

Moreover, ${\bf{D}}_{r;k}^{\left( n \right)} = {\bf{D}}_{r;k - 1}^{\left( n \right)} + \beta _{k - 1}({{\cal R}}{{{\cal Q}}_r}{\bf{U}}_k^{\left( n \right)} - {\bf{Z}}_{r;k}^{\left( n \right)})$ and \eqref{eq:ADMM_5} imply
\begin{equation}\label{theorem_1_2}
\begin{aligned}
& {\beta _k}\sum\limits_{r = 1}^{\hat R} {{{\cal Q}}_r^*{{{\cal R}}^*}{{\cal R}}{{{\cal Q}}_r}\big( {{\bf{U}}_{k + 1}^{\left( n \right)} - {\bf{U}}_k^{\left( n \right)}} \big)}  \\
= & \lambda {{{\cal A}}_k^{\left( n \right)*}}\left( {{{{\cal P}}_{{\Omega ^{(n)}}}}({{\bf{Y}}_{(n)}})} \right) - \lambda {{\cal A}}_k^{\left( n \right)*}{{\cal A}}_k^{\left( n \right)}{\bf{U}}_{k + 1}^{\left( n \right)} \\
& +{\beta _k}\sum\limits_{r = 1}^{\hat R} {{{\cal Q}}_r^*{{{\cal R}}^*}\big( {{\bf{Z}}_{r;k}^{(n)} - {{({\beta _k})}^{ - 1}}{\bf{D}}_{r;k}^{(n)} - {\cal R}{{{\cal Q}}_r}{\bf{U}}_k^{\left( n \right)}} \big)} \\
= &\lambda {{{\cal A}}_k^{\left( n \right)*}}\left( {{{{\cal P}}_{{\Omega ^{(n)}}}}({{\bf{Y}}_{(n)}})} \right) - \lambda {{\cal A}}_k^{\left( n \right)*}{{\cal A}}_k^{\left( n \right)}{\bf{U}}_{k + 1}^{\left( n \right)} \\
& -\sum\limits_{r = 1}^{\hat R} {{{\cal Q}}_r^*{{{\cal R}}^*}\big( {\left( {\rho  + 1} \right){\bf{D}}_{r;k}^{(n)} - \rho {\bf{D}}_{r;k - 1}^{(n)}} \big)}.
\end{aligned}
\end{equation}

According to \eqref{eq:OP_3}, we have
\begin{equation}\label{theorem_1_3}
\begin{aligned}
& {\bf{C}} \circledast \big( {{\bf{U}}_{k + 1}^{\left( n \right)} - {\bf{U}}_k^{\left( n \right)}} \big) \\
= & \frac{1}{{{\beta _k}}}\big( {\lambda {{{\cal A}}_k^{\left( n \right)*}}\left( {{{{\cal P}}_{{\Omega ^{(n)}}}}({{\bf{Y}}_{(n)}})} \right) } - { \lambda {{\cal A}}_k^{\left( n \right)*}{{\cal A}}_k^{\left( n \right)}{\bf{U}}_{k + 1}^{\left( n \right)} }\\
& -{\sum\limits_{r = 1}^{\hat R} {{{\cal Q}}_r^*{{{\cal R}}^*}\big( {\left( {\rho  + 1} \right){\bf{D}}_{r;k}^{(n)} - \rho {\bf{D}}_{r;k - 1}^{(n)}} \big)} } \big) = \frac{{{{\bf{L}}_k}}}{{{\beta _k}}},
\end{aligned}
\end{equation}
where ${\bf{C}}$ is a constant matrix and $ \circledast $ denotes Hadamard product. Since ${{\bf{L}}_k}$ is upper bounded, ${\big\| {{\bf{U}}_{k + 1}^{\left( n \right)} - {\bf{U}}_k^{\left( n \right)}} \big\|_F} = O\big( {{{\left( {{\beta _k}} \right)}^{ - 1}}} \big)$. We further have
\begin{equation}\label{theorem_1_4}
\begin{aligned}
& {\big\| {{\bf{C}} \circledast \big( {{\bf{U}}_m^{\left( n \right)} - {\bf{U}}_k^{\left( n \right)}} \big)} \big\|_F} \le {\big\| {{\bf{C}} \circledast \big( {{\bf{U}}_m^{\left( n \right)} - {\bf{U}}_{m - 1}^{\left( n \right)}} \big)} \big\|_F} \\
& +{\big\| {{\bf{C}} \circledast \big( {{\bf{U}}_{m - 1}^{\left( n \right)} - {\bf{U}}_{m - 2}^{\left( n \right)}} \big)} \big\|_F} +  \ldots  + \\
& {\big\| {{\bf{C}} \circledast \big( {{\bf{U}}_{k + 1}^{\left( n \right)} - {\bf{U}}_k^{\left( n \right)}} \big)} \big\|_F}\\
= & \frac{{{{\big\| {{{\bf{L}}_{m - 1}}} \big\|}_F}}}{{{\beta _{m - 1}}}} + \frac{{{{\big\| {{{\bf{L}}_{m - 2}}} \big\|}_F}}}{{{\beta _{m - 2}}}} +  \ldots  + \frac{{{{\big\| {{{\bf{L}}_k}} \big\|}_F}}}{{{\beta _k}}} \le \\
& \frac{\delta }{{{\beta _k}}}\big( {\frac{1}{{{\rho _{m - k - 1}}}} + \frac{1}{{{\rho _{m - k - 2}}}} +  \ldots  + 1} \big) < \frac{{\delta \rho }}{{{\beta _k}\left( {1 - \rho } \right)}},
\end{aligned}
\end{equation}
where $\delta  = \max \left\{ {{{\left\| {{{\bf{L}}_{m - 1}}} \right\|}_F}, {{\left\| {{{\bf{L}}_{m - 2}}} \right\|}_F}, \ldots ,{{\left\| {{{\bf{L}}_k}} \right\|}_F}} \right\}$. Since $\frac{{\delta \rho }}{{{\beta _k}\left( {1 - \rho } \right)}} \to 0$, $\{ {\bf{U}}_k^{\left( n \right)}\} $ is a Cauchy sequence for all $n \in \{ 1, \ldots ,N\}$, i.e., $\{\mathcal{U}_k\}$ is a Cauchy sequence.

Similarly, $\{ {\bf{Z}}_{r;k}^{\left( n \right)}\} $, $r = 1, \ldots ,\hat R$, $n = 1, \ldots ,N$, also is a Cauchy sequence.

\vspace{0.4cm}
% proof of theorem 2
\noindent \textit{Proof of Theorem 2}:
Let $(\tilde{\bf {U}}_k^{\left( 1 \right)} \cdots ,\tilde{\bf{ U}}_k^{\left( N \right)})$ be a stationary point of \eqref{eq:proposednew}, then the KKT conditions of \eqref{eq:proposednew} are
\begin{equation}\label{theorem_2_1}
\begin{aligned}
{\bf{0}} \in &\sum\limits_{r = 1}^{\hat R} {{{\cal Q}}_r^*{{{\cal R}}^*}\partial {{\big\| {{{\cal R}}{{{\cal Q}}_r}{{\tilde{{\bf{U }}}}^{\left( n \right)}}} \big\|}_*}}  \\
& +\lambda {{\tilde{\cal A}}}^{\left( n \right)*}{{\tilde{\cal A}}}^{\left( n \right)}{\tilde{\bf{U}}^{\left( n \right)}} - \lambda {{{\tilde{\cal A}}}^{\left( n \right)*}}\left( {{{{\cal P}}_{{\Omega ^{(n)}}}}({{\bf{Y}}_{(n)}})} \right), \\
\end{aligned}
\end{equation}
for all $n \in \{ 1, \ldots ,N\}$, where ${{\tilde{\cal A}}}^{\left( n \right)}\left( {\bf{X}} \right) \buildrel \Delta \over = {{{\cal P}}_{{\Omega ^{\left( n \right)}}}}\big( {\bf{X}}\tilde{\bf{G}}^{\left( n \right)} \big)$ and ${\tilde{\bf{G}}}^{\left( n \right)} = ({\tilde{\bf{U}}}^{\left( N \right)} \odot  \cdots  \odot {\tilde{\bf{U}}}^{\left( {n + 1} \right)} \odot {\tilde{\bf{U}}}^{\left( {n - 1} \right)} \odot  \cdots  \odot {\tilde{\bf{U}}}^{\left( 1 \right)})^T$.

According to Algorithm \ref{alg:1}, the first-order optimal condition of \eqref{eq:ADMM_4} at the (\textit{k}+1)-th iteration is
\begin{equation}\label{theorem_2_2}
\begin{aligned}
{\bf{0}} &=\lambda {{\cal A}}_k^{\left( n \right)*}{{\cal A}}_k^{\left( n \right)}{{\bf{U}}_{k+1}^{\left( n \right)}} -\lambda {{\cal A}}_k^{\left( n \right)*}\left( {{{{\cal P}}_{{\Omega ^{(n)}}}}({{\bf{Y}}_{(n)}})} \right) \\
&+ {\beta _k}\sum\limits_{r = 1}^{\hat R} {{\cal Q}_r^*{{\cal R}^*}({\cal R}{{\cal Q}_r}{\bf{U}_{k+1}^{\left( n \right)}}-{\bf{Z}}_{r;k}^{(n)}+{({\beta _k})}^{ - 1}{\bf{D}}_{r;k}^{(n)})}.  \\
\end{aligned}
\end{equation}

The first-order optimal condition of the problem \eqref{eq:ADMM_7} is
$$
{\bf{0}} \in \partial {\big\| {{\bf{Z}}_{r;k + 1}^{(n)}} \big\|_*} + \beta _k({\bf{Z}}_{r;k + 1}^{(n)} - {{\cal R}}{{{\cal Q}}_r}{\bf{U}}_{k + 1}^{\left( n \right)}) - {\bf{D}}_{r;k}^{(n)},
$$
implying that
\begin{equation}\label{theorem_2_3}
{\bf{0}}\in\partial{\big\| {{\bf{Z}}_{r;k + 1}^{(n)}} \big\|_*}-{\bf{D}}_{r;k+1}^{(n)}.
\end{equation}

Since $\{ {\bf{U}}_k^{\left( n \right)}\} $ and $\{ {\bf{Z}}_{r;k}^{\left( n \right)}\}$ are all Cauchy sequences, $\{ {\bf{U}}_\infty ^{\left( n \right)}\} $ and $\{ {\bf{Z}}_{r;\infty }^{\left( n \right)}\} $ are their limit points, respectively. We also have ${{\cal R}}{{{\cal Q}}_r}{{\bf{U}}_\infty^{\left( n \right)} } = {\bf{Z}}_{r;\infty }^{\left( n \right)}$ according to \eqref{theorem_1_1}. By \eqref{theorem_2_2} and \eqref{theorem_2_3}, if ${\lim _{k \to \infty }}\big\| {{\bf{D}}_{r;k + 1}^{(n)} - {\bf{D}}_{r;k}^{(n)}} \big\|_F=0$ for all $r = 1, \ldots ,\hat R$ and $n = 1, \ldots ,N$, we have
\begin{equation}\label{theorem_2_4}
\begin{aligned}
{\bf{0}} \in &\sum\limits_{r = 1}^{\hat R} {{{\cal Q}}_r^*{{{\cal R}}^*}\partial {{\big\| {{{\cal R}}{{{\cal Q}}_r}{\bf{U}}_\infty ^{\left( n \right)}} \big\|}_*}}  \\
& +\lambda {{\cal A}}_\infty^{\left( n \right)*}{{\cal A}}_\infty^{\left( n \right)}{\bf{U}}_\infty ^{\left( n \right)} - \lambda {{{\cal A}}_\infty^{\left( n \right)*}}\left( {{{{\cal P}}_{{\Omega ^{(n)}}}}({{\bf{Y}}_{(n)}})} \right).
\end{aligned}
\end{equation}
where ${{\cal A}}_\infty^{\left( n \right)}\left( {\bf{X}} \right) \buildrel \Delta \over = {{{\cal P}}_{{\Omega ^{\left( n \right)}}}}\left( {{\bf{XG}}_\infty^{\left( n \right)}} \right)$ and ${\bf{G}}_\infty^{\left( n \right)} = ({\bf{U}}_\infty^{\left( N \right)} \odot  \cdots  \odot {\bf{U}}_\infty^{\left( {n + 1} \right)} \odot {\bf{U}}_\infty^{\left( {n - 1} \right)} \odot  \cdots  \odot {\bf{U}}_\infty^{\left( 1 \right)})^T$

Therefore the limit of the sequence $\{\mathcal{U}_k\}$ satisfies the KKT conditions of the problem \eqref{eq:proposednew}.

\end{appendix}

% ================================================================
% -------------------------- Reference ---------------------------
% ================================================================
\ifCLASSOPTIONcaptionsoff
  \newpage
\fi

\bibliographystyle{IEEEtran}

\end{document}